\documentclass{article}

\usepackage{iclr2027_conference,times}

\usepackage[T1]{fontenc}
\usepackage[utf8]{inputenc}
\usepackage{microtype}
\usepackage{inconsolata}
\usepackage{helvet}                      % [CoLab] Helvetica clone for the header

\usepackage{amsmath}
\usepackage{graphicx}
\usepackage{wrapfig}
\usepackage{float}
\usepackage{booktabs}
\usepackage{url}

\usepackage{xcolor}                      % [CoLab]
\usepackage{titlesec}                    % [CoLab]
\usepackage{chessfss}                    % [CoLab] affiliation symbols

\graphicspath{{figures/}{media/}}        % [CoLab] so the logo resolves

\definecolor{headergray}{HTML}{293138}   % [CoLab]

\titleformat{\section}
  {\Large\bfseries}{\thesection}{0.6em}{}
\titleformat{\subsection}
  {\large\bfseries}{\thesubsection}{0.6em}{}
\titleformat{\subsubsection}
  {\normalsize\bfseries}{\thesubsubsection}{0.6em}{}

\usepackage[colorinlistoftodos,disable]{todonotes}
\extrafloats{100} %% headroom for the 19 real figures/tables, several of which land in one appendix section
\usepackage{xspace}
\newcommand{\cready}[1]{} % a hidden comment to remind us for the final version (e.g. \cready{add a link for github})

\usepackage{enumitem}
  \setlist{itemsep=0.5pt, topsep=0pt,leftmargin=1.4em,labelsep=0.5em}

\usepackage{hyperref} 
\usepackage{xcolor}
\hypersetup{
    colorlinks,
    linkcolor={red!50!black},
    citecolor={blue!50!black},
    urlcolor={blue!80!black}
}
\usepackage{amsmath}
\usepackage[capitalize,nameinlink]{cleveref}
\AddToHook{cmd/appendix/before}{\crefalias{section}{appendix}}
\Crefname{equation}{Eq.}{Eqs.}
\Crefname{figure}{Fig.}{Figs.}
\Crefname{tabular}{Tab.}{Tabs.}

\usepackage{etoolbox} % for \pretocmd
\makeatletter
\pretocmd{\appendix}{%
  \@addtoreset{figure}{section}%
  \@addtoreset{table}{section}%
}{}{}
\makeatother

\newcommand{\fkd}{FKD\xspace}
\newcommand{\hfkd}{HFKD\xspace}

\DeclareRobustCommand{\ceq}[1][]{%
  \ensuremath{\text{CL\textsubscript{eq}}\if\relax\detokenize{#1}\relax\else_{#1}\fi}%
  \ifmmode\else\xspace\fi}
\newcommand{\enar}{\mbox{English--Arabic}\xspace}
\newcommand{\enru}{\mbox{English--Russian}\xspace}
\newcommand{\enen}{\mbox{English\textsubscript{1}--English\textsubscript{2}}\xspace}

\title{Why Pretraining Fails to Share Cross-Lingual Knowledge}

\iclrfinalcopy

\begin{document}

% [CoLab] Drop the "Published as a conference paper at ICLR 2027"
% running head. Comment out if you are submitting rather than
% posting a preprint.
\pagestyle{plain}
\thispagestyle{plain}

% ------------------------------------------------------------------
% CoLab header bar
% ------------------------------------------------------------------
\noindent
\raisebox{-0.22\height}{%
  \includegraphics[height=1.6em]{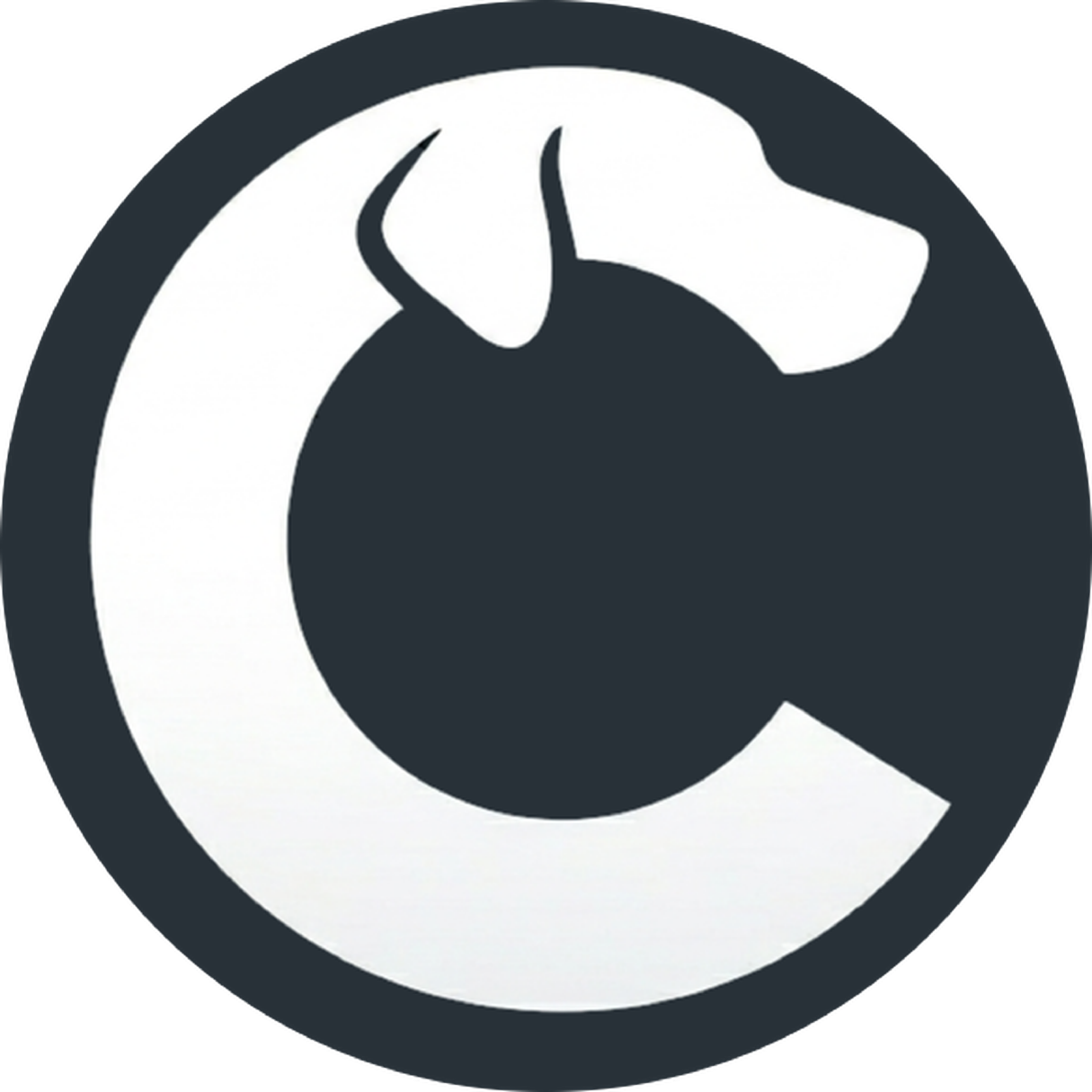}%
}%
\hspace{0.4em}%
{\color{headergray}\sffamily\Large\bfseries CoLab Lab}%
\hfill
{\color{headergray}\small 2026-09-16}

\vspace{0.15em}

\noindent
{\color{headergray}\rule{\textwidth}{0.5pt}}

\vspace{1.4em}

% ------------------------------------------------------------------
% Title
% ------------------------------------------------------------------
\noindent
{\fontsize{22}{26}\selectfont\bfseries
Why Pretraining Fails to Share Cross-Lingual Knowledge
\par}

\vspace{0.9em}

% ------------------------------------------------------------------
% Authors
% ------------------------------------------------------------------
\noindent
{\fontsize{10.5}{13}\selectfont
\mbox{\textbf{Adam Gaber}$^{1,\dagger}$},
\mbox{\textbf{Uriel Dolev}$^{2}$},
\mbox{\textbf{Elisabeth Fittschen}$^{3}$},
\mbox{\textbf{Bobby Cheng}$^{4}$},
\mbox{\textbf{Yuval Marton}$^{5}$},
\mbox{\textbf{Leshem Choshen}$^{1, 6,7,\ast}$}
\par
}

\vspace{0.3em}

% ------------------------------------------------------------------
% Affiliations
% ------------------------------------------------------------------
\noindent
{\fontsize{9}{11}\selectfont
$^{1}$Weizmann Institute of Science,
$^{2}$Bar-Ilan University,
$^{3}$Johns Hopkins University,
$^{4}$A*STAR,
$^{5}$University of Washington,
$^{6}$MIT,
$^{7}$MIT-IBM Watson AI Lab
\par
}

\vspace{1.8em}

% ------------------------------------------------------------------
% Abstract
% ------------------------------------------------------------------
\begin{center}
\begin{minipage}{0.90\textwidth}
\small
Large Language Models (LLMs) have made remarkable progress in the processing and
modeling of many languages. Yet, unlike human multilinguals, they exhibit
surprisingly limited cross-lingual knowledge transfer. While this limitation is
well documented, its origins during multilingual training remain unclear. We
pretrain 360M- and 7B-parameter LLMs and show that poor cross-lingual knowledge
generalization emerges during pretraining and persists under standard
interventions. To isolate its cause, we employ a controlled bilingual pretraining
setting using two copies of the same language, sharing identical text and token
segmentation, but mapped to disjoint token spaces. We find that disjoint tokens
alone are enough to induce knowledge compartmentalization, even between identical
copies of the same language, establishing disjoint token spaces as a fundamental
barrier to cross-lingual knowledge generalization. Guided by this understanding,
we suggest mapping languages into a shared token space by simple word-wise
translation and find it substantially improves cross-lingual knowledge
generalization, recovering up to 12.6\% of native-language learning efficiency
--- 14$\times$ the baseline.
\end{minipage}
\end{center}

\vspace{1.6em}

% ------------------------------------------------------------------
% Corresponding-author footnotes
% ------------------------------------------------------------------
\begingroup
\renewcommand{\thefootnote}{\fnsymbol{footnote}}

\footnotetext[2]{Corresponding author:
  \href{mailto:adam.gaber@weizmann.ac.il}{\nolinkurl{adam.gaber@weizmann.ac.il}}}

\footnotetext[1]{Corresponding author:
  \href{mailto:leshem.choshen@mail.huji.ac.il}{\nolinkurl{leshem.choshen@weizmann.ac.il}}}

\footnotetext[3]{\label{fn:repo}Code publicly available
  \url{https://github.com/AdamJaber03/torchtitan-mulitlingual}}

\endgroup

% ==================================================================
% Everything below here is your existing paper, unchanged.
% Delete the old \maketitle and \begin{abstract}...\end{abstract}.
% ==================================================================

\label{sec:intro}
\begin{figure}[ht]
    \centering
    \includegraphics[width=1.0\linewidth]{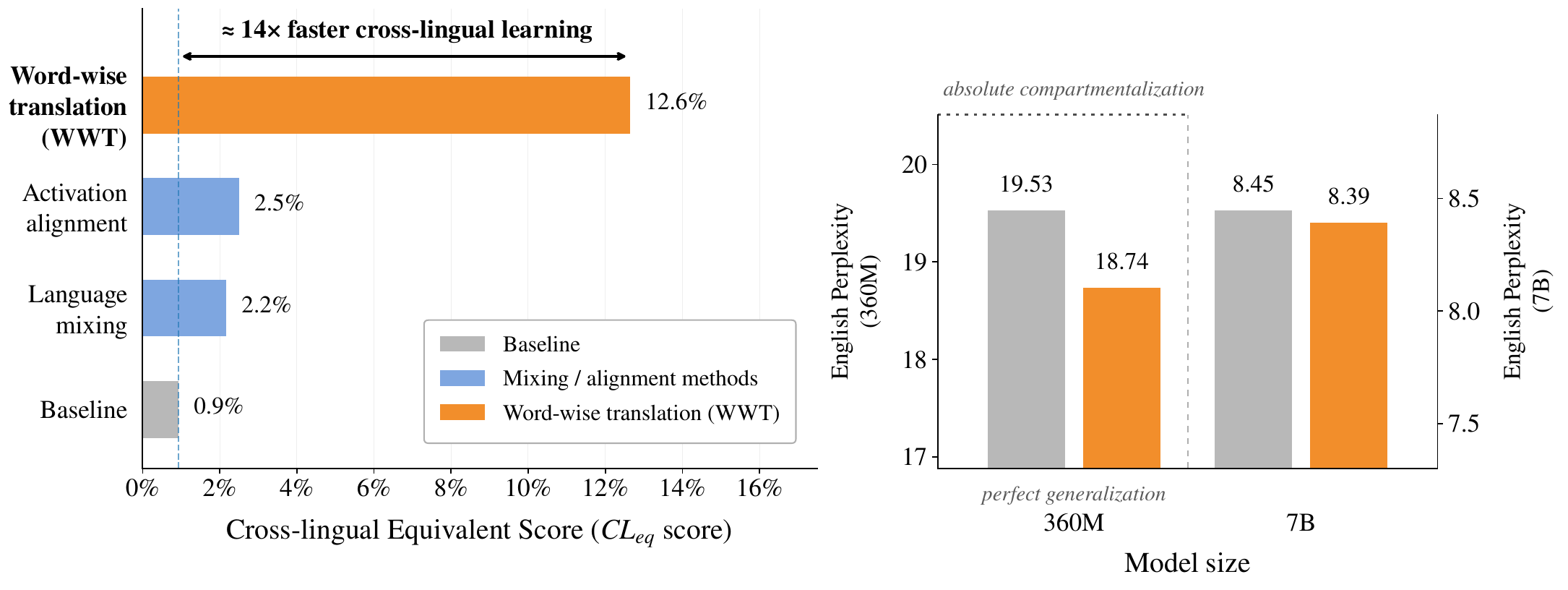}
    \caption{
    Merging token spaces via WWT yields substantial gains in cross-lingual knowledge generalization. Left: \ceq for \enar\ pretraining under various methods (colors), measuring cross-lingual knowledge learning efficiency at 360M scale. Right: left panel shows English perplexity at 360M; upper bound is a model trained on the English half alone, the lower bound is a model trained on twice as much English.  Right panel shows perplexity at 7B, bounds are proportional to 360M.
    % \lc{in all captions, we can state a bottom line shortly at the beginning, but the main body should describe the figure not the results of the type:(bottom line sentence. Depicts a comparison across methods (colors) in their Cross-lingual Equivalence (top) and perplexity (bottom). more or less content is style plus what really needs explaining, but the caption can only make a short statement most of that is for that text}
    }
    \label{fig:enArCeq}
    \vspace{-0.2cm} % Adjust this value (e.g., -1cm, -5mm) until the spacing looks right

\end{figure}
\section{Introduction} 
% \lc{before submitting make sure no figures make the annoying end of page bug, no comments no ?? check warnings and compilation errors, make sure you have all authors and affiliations and links in the right way (anon or not depending on where it goes.}
% P1
Modern LLMs exhibit impressive multilingual capabilities, yet unlike humans, they transfer surprisingly little knowledge across languages. Understanding this limitation is important for building truly multilingual models.

% P2
Prior work has documented poor cross-lingual generalization, characterizing it through model capacity and sample efficiency, and exploring factors such as tokenization, language imbalance, and activation alignment \citep{chang2023multilingualitycurselanguagemodeling, schäfer2024rolelanguageimbalancecrosslingual, howe2026languagemodelsstrugglecompartmentalization, wu-dredze-2020-explicit, li-etal-2024-prealign}. Recent work extends this generalization failure into the knowledge domain \citep{Qi_2023, ifergan2024beneathsurfaceconsistencyexploring, goldman2025eclekticnovelchallengeset} and tries to reduce it \citep{huang-etal-2023-languages,  diskind2026crosslingualexplorationparametricknowledge, bandarkar2026largereasoningmodelsstruggle, vonrad2026improvingcrosslingualfactualrecall}. However, these observations and fixes are made post-hoc on fully trained models, which go through a sequence of training stages including pretraining, instruction tuning, and preference optimization, leaving open where this knowledge generalization failure, which we refer to as compartmentalization, originates.

% a complex argument but maybe worth noting that while post-hoc solutions enable access to knowledge previously unaccessble, they still dont solve the inner-model compartmentalization of this knowledge. that is knowing a in langA and b in langB you can access both a and b with post-hoc solutions but they probably wont enable access to c which arises from understanding a and b together.

% P3
We investigate compartmentalization directly at the pretraining stage. However, causal analysis in this setting is complicated by two factors: First, pretraining corpora are vast, and tracking knowledge exposure across languages is intractable. Second, natural languages simultaneously differ in vocabulary, scripts, syntax, frequency distributions and semantics, hindering controlled experiments.

% P4
We address these challenges through a two-part methodology. First, to reliably trace knowledge exposure, we introduce a controlled pretraining environment utilizing the injection of fictive knowledge (\S\ref{sec:diagnosis_method}). Second, to strip away linguistic confounds, we evaluate generalization not only on natural bilingual pairs but also on structurally identical, cloned language copies, that differ in nothing but their token spaces (\S\ref{sec:tokensRcause:en1en2}). Using this combined framework, we pretrain 360M- and 7B-parameter models to measure and isolate the mechanisms of cross-lingual knowledge generalization.

% P5
Our controlled measurements reveal that knowledge \textbf{compartmentalization already emerges during pretraining} (\S\ref{sec:pretraining_flawed}). We further identify \textbf{disjoint token spaces as a fundamental barrier} to the knowledge generalization mechanism that persists across model scales. Two languages represented in disjoint regions of the model's vocabulary, regardless of token segmentation, hinder sharing (\S\ref{sec:tokensRcause}).

% P6
Guided by our newfound understanding of this mechanism, we suggest a word-wise translation (WWT) intervention to efficiently unify the token space of vocabulary-disjoint languages (\S\ref{sec:TrAr:TrAr}). This intervention proves beneficial, substantially improving cross-lingual knowledge generalization while providing gains in general language modeling perplexity (Fig.~\ref{fig:enArCeq}). This practical success provides a mechanistic validation of our findings on tokenization's role in knowledge generalization (\S\ref{sec:TrAr:res}).

% Contributions...
\section{Methods: Measuring Knowledge Generalization}
% To diagnose cross-lingual knowledge transfer, we first establish our toolset.
\label{sec:diagnosis_method}
\subsection{Controlling knowledge exposure - Injecting fictive facts}
\label{sec:injecting}
To measure if and to what extent exposure to a fact during training in one language yields knowledge of that fact in another language, we must know how often the fact has been observed in each language. Tracing fact exposures in the vast pretraining corpora is often intractable~\citep{kandpal2023largelanguagemodelsstruggle, elazar2024whatsbigdata}. We bypass this problem by injecting novel fictive knowledge, absent from the underlying pretraining data, with precise control over the number and language of exposures to each fact. 
% This enables us to measure both the exposure required for acquisition and the extent to which knowledge generalizes across languages.

% To measure if exposure in one language implies knowing in a second, we first need to determine the amount of exposure in each language. Usually, tracing this knowledge exposure is complicated as the pretraining data is vast \citep{kandpal2023largelanguagemodelsstruggle, elazar2024whatsbigdata}. In our case, however, we bypass this complexity by controlling the training data and injecting facts that are certainly not already in it, determining the precise number of exposures in each language to these facts and using them to evaluate performance. Specifically, we generate fictive facts about entities and an associated attribute (e.g., "Cultramive is a company that sells silver-plated bicycles"). \lc{we mention somewhere in the results the byproduct of knowing how much exposure one needs, right?}

Our Fictional Knowledge Dataset~(\fkd) contains 2,048 facts about synthetic entities and associated attributes (e.g., ``Cultramive is a company that sells silver-plated bicycles''). Each fact is expanded into a per-language set of diverse natural documents that paraphrase the same underlying knowledge --- used for injection during pretraining --- and held-out multiple-choice questions (MCQs) for evaluation. We provide full details on \fkd, including the generation pipeline and examples in App.~\ref{app:fictive}. Since prior work indicates that cross-lingual generalization mechanisms can behave differently for high- and low-frequency tokens \citep{schäfer2024rolelanguageimbalancecrosslingual, feng-etal-2025-word}, we replicate our main experiments on a fictive-persona dataset built around high-frequency human names (App.~\ref{app:hfed}).

We validate \fkd by pretraining monolingual models with entity facts injected at different rates (App.~\ref{app:monolingual}). At 0-injection, accuracy on the MCQs is around chance, injecting fact paraphrasings into pretraining increases accuracy in a log-like manner, reaching substantial accuracy across all languages; \fkd is learnable and controllable.

\subsection{Evaluating Cross-lingual knowledge generalization -- \ceq Score}

While fictive knowledge allows us to control how it is introduced across languages, we're still missing a direct measure, given a bilingual model, that quantifies how well it generalizes factual knowledge between languages $A$ and $B$. To this end we introduce the Cross-Lingual equivalence (\ceq) score.

% Let \(V_A(B)\) denote the value of one exposure to a fact in language \(B\) for recalling that fact in language \(A\), and let \(V_A(A)\) denote the value of one exposure in the evaluation language itself. We define the directional \ceq score as
% \[
% \ceq[B \rightarrow A]
% =
% 100 \times \frac{V_A(B)}{V_A(A)}.
% \]
% \adam{since we now present the log-like acquisition curve before this, the "value of one exposure" becomes not well defined as it is dependent on how many exposure have already been. thoughts?}
% Thus, 
\(\ceq[B \rightarrow A]\) measures how valuable an exposure to a fact in language \(B\) is for recalling that fact in language \(A\), relative to a native exposure in \(A\) itself. A score of \(100\%\) indicates that exposure in \(B\) is as effective as exposure in \(A\), corresponding to perfect cross-lingual equivalence. A score of \(0\%\), in contrast, indicates that exposure in \(B\) provides no benefit for recall in \(A\), corresponding to complete knowledge compartmentalization across the two languages.

% As cross-lingual transfer may be asymmetric \citep{}, we compute \ceq separately in both directions. 
When reporting a single score for a given bilingual model trained on languages $A$ and $B$, we use the bidirectional average:
% \[ \ceq_{A\leftrightarrow B} = \frac{1}{2} \left( \ceq[A \rightarrow B] + \ceq[B \rightarrow A] \right). \]
\[ \ceq = \frac{1}{2} \left( \ceq[A \rightarrow B] + \ceq[B \rightarrow A] \right). \]

% The directional \ceq score, $\ceq[B \rightarrow A]$, measures how much an exposure in the other language is better than an exposure in the same language tested. \lc{this can be a formula right? relaly formal, given bla the exposure in lang $A$...} We  value of one exposure to a fact in language $B$ for recalling that fact in language $A$, expressed as a percentage of the value of one native exposure in language $A$. 
% A score of 100\% indicates perfect cross-lingual knowledge generalization, whereas a score of 0\% indicates absolute knowledge compartmentalization.

% When reporting a single \ceq score, we represent the average of both directions: $\frac{1}{2}(\ceq[A \rightarrow B] + \ceq[B \rightarrow A])$.\lc{Should we then also give it a different symbol? Like just ceq without A->B. Something like, We report the general transfer ability as ceq, the average of... (I didnt change because need to ensure we are consistent with the symbols in all references and graphs}

Crucially, our \ceq\ definition normalizes cross-lingual gains by native ones. This makes our metric invariant to differences in baseline model competence (raw accuracy) that may arise from scale, training recipe, or data quality; a model that learns all facts better raises foreign and native exposure value together, leaving their ratio unchanged. This separates gains in target-language ability from gains in transfer, which measurements reported in absolute accuracy terms conflate.

\paragraph{Estimating \ceq score} 
\label{sec:ceq}
Controlling \fkd exposure enables a direct estimate of a model's \ceq{} score. For a fictive fact $f$ in \fkd, let $N_A(f)$ and $N_B(f)$ denote the number of exposures to it during pretraining in languages $A$ and $B$ respectively, and let $\mathrm{acc}_A(f)$ denote trained model accuracy on $f$'s language-$A$ evaluation questions. We fit a linear regression (OLS) over all facts,
\begin{equation}
\label{eq:ols}
    \mathrm{acc}_A(f) = \alpha + \beta_A\, N_A(f) + \beta_B\, N_B(f) + \varepsilon,
\end{equation}
$\beta_A$ captures the accuracy gained per native exposure, $\beta_B$ captures the accuracy gained per cross-lingual exposure, and the score is derived from their ratio:
\begin{equation}
\label{eq:ceq}
    \ceq[B \rightarrow A] = 100 \times \frac{\beta_B}{\beta_A}.
\end{equation}

Two features of this estimate warrant note. First, the OLS regression provides a linear approximation to the relationship between exposure and accuracy (\S\ref{sec:injecting}); we experiment with an alternative curve-aware estimator and find the simple linear form more informative. Second, sampling and evaluation noise can occasionally pull estimates slightly below zero. Negative values should be read as substantial or absolute compartmentalization rather than negative transfer. App.~\ref{app:ceq} covers these topics in detail, and reports significance measures confirming our headline comparisons are statistically robust.

\section{Experimental Setup}

Our experiments follow a common template. We pretrain bilingual models from scratch, injecting \fkd fictive facts at varying cross-lingual rates, and estimate \ceq score based on their acquisition. We vary isolated aspects of the bilingual setup (architecture, tokenization, and interventions) and compare the resulting scores to reason about cross-lingual knowledge generalization. Each of the presented \ceq results requires a full model pretraining run corresponding to its unique condition.

\subsection{Models and data}
We evaluate two scales: a 360M-parameter model following the SmolLM2 architecture \citep{allal2025smollm2smolgoesbig}, and a 7B-parameter adaptation of Llama-3 \citep{grattafiori2024llama3herdmodels}. Both use a BPE tokenizer with a 65{,}536-token vocabulary. We train one tokenizer per language pair on a 50/50 split of 7M documents from each of its two languages, and share it across the matching runs. Models whose perplexities we compare therefore read text under identical segmentation.
% doubled to 131{,}072 for $\text{English}_1$--$\text{English}_2$ experiments to accommodate two disjoint copies. 
Models are trained to approximately Chinchilla-optimal token budgets \citep{hoffmann2022trainingcomputeoptimallargelanguage} using TorchTitan \citep{liang2025torchtitanonestoppytorchnative}. Architectures, hyperparameters, schedules, and hardware are detailed in App.~\ref{app:training}.

We pretrain on \textit{FineWeb-edu} for English \citep{penedo2024finewebdatasetsdecantingweb}, \textit{FineWeb-edu-ar} for Arabic \citep{alrashed2024finewebeduarmachinetranslatedcorpussupport}, a machine translation of FineWeb-edu, and \textit{FineWeb2-HQ} for Russian \citep{messmer2026enhancingmultilingualllmpretraining}, native web text rather than translation. Our bilingual experiments use a 50/50 token budget split between languages. As the English and Arabic corpora are translations of one another, we enforce a strict document-disjoint split to ensure models do not see parallel documents across languages.
%\efi{ Minor style note: I think em-dashes have gotten a bad reputation recently due to how common ai generated texts produce them. Might be worth swapping some for commas, parentheses, or sentence boundaries, though entirely optional.}

\subsection{Evaluation}
\label{sec:expsetup:eval}
% \subsection{Calculating \ceq}
To calculate the \ceq metric, we split \fkd into 16 groups of 128 fictive entity facts. Each of the 16 groups has all its fictive facts injected at a uniform $N_A$ and $N_B$ chosen from the cross-product of [0,20,100,1000] with itself. This range was selected to be denser on the critical ranges of the fact learning curve (App.~\ref{app:monolingual}). Fact paraphrasing documents are injected stochastically into pretraining at a per-language probability matching the target exposure counts $N_A$ and $N_B$ (App.\ref{app:training:inj}).

We use the LM-eval-harness \citep{eval-harness} evaluation protocol along with the accompanying \fkd per-language evaluation MCQs (generated separately from the injection documents and never present in training) to quantify the model's fact acquisition across the different languages: $\mathrm{acc}_A$ and $\mathrm{acc}_B$. Specifically, we use acc-norm scoring from LM-eval-harness to rectify for possible variance in choice length. With these values at hand, we fit the regression of Eq.~\ref{eq:ols} across all facts and take \ceq as the ratio of $\beta_A$ and $\beta_B$ (\S\ref{sec:ceq}). We report fitted coefficients for representative runs in App.~\ref{app:ceq}

\section{Knowledge Compartmentalization Originates in Pretraining}
\label{sec:pretraining_flawed}

\subsection{Compartmentalization emerges in standard bilingual pretraining}
Applying the above-mentioned tools, we show that current pretraining practices strongly compartmentalize knowledge. We pretrain a 360M-parameter \enar\ LLM, injecting fictive facts at controlled cross-lingual rates (\S\ref{sec:injecting}). With the trained bilingual model at hand, we measure fictive fact recall and deduce the relative value of foreign exposure to knowledge --- the \ceq score (\S\ref{sec:ceq}). The pretrained \enar\ model shows minuscule knowledge generalization, with a \ceq score of 0.9\% (Fig.~\ref{fig:enArCeq}). Severe knowledge compartmentalization is also apparent in an \enru\ setup with $\ceq=1.9\%$ and at the 7B parameter scale with $\ceq=5.9\%$ (Tab.~\ref{tab:wwt_results}), suggesting knowledge compartmentalization in pretraining is not an artifact of our specific language pairing or scale.

Analyzing per-language perplexity, we find that a bilingual setup is slightly better than pretraining on either of the monolingual halves of the corpus alone (Fig.~\ref{fig:enArCeq}), consistent with previous works demonstrating positive transfer \citep{conneau2020unsupervisedcrosslingualrepresentationlearning}. However, it appears that whatever is driving this gain in language modeling, it is not knowledge generalization.

\subsection{Standard interventions fail}
\label{sec:pretrainnig_flawed:interventions}

% \paragraph{Standard interventions fail} 
As bilingual pretraining naturally compartmentalizes knowledge, we experiment with pretraining interventions suggested for related purposes to resolve it, and find that they largely fail. In particular, we experiment with two types of pretraining interventions in the \enar\ setting at 360M-parameter scale: language mixing and activation alignment (App.~\ref{app:interventions}).

Language mixing introduces multiple languages within a single input sequence, encouraging the model to utilize cross-lingual information. Our language mixing variant of choice is pretraining code-switching (named after linguistic code-switching), which randomly substitutes words with their cross-lingual counterparts \citep[e.g.][]{zeng2026bringingbilingualbabylminvestigating}. We use a comprehensive word-level dictionary for the substitution operation, and sweep both the fraction of mixed documents and the within-document substitution rate, finding no substantial benefit, with a \ceq score of 2.2\%.

Activation alignment uses auxiliary losses to align pooled activations of semantically equivalent cross-lingual units. We experiment with InfoNCE \citep{oord2019representationlearningcontrastivepredictive, chi2021infoxlminformationtheoreticframeworkcrosslingual} and L2 loss estimating sequence matches by chunking parallel documents. Across various chunk sizes we find activation alignment offers no help in resolving compartmentalization: \ceq score of 2.5\%.

Taken together, our results establish that knowledge compartmentalization emerges already in pretraining, persists under standard interventions, and is attenuated but not resolved by scale, suggesting that a fundamental solution should address the breakdown at this stage. What causes this breakdown remains open: natural language pairs differ in script, grammar, segmentation, token vocabulary, and data distribution at once, so these measurements alone cannot say which difference is responsible.

\section{Disjoint Token Space is a Root Cause of Compartmentalization}
\label{sec:tokensRcause}

\begin{wrapfigure}{r}{0.55\textwidth} % "r" aligns right, reserving 55% of the page width
\centering
    \includegraphics[width=1.0\linewidth]{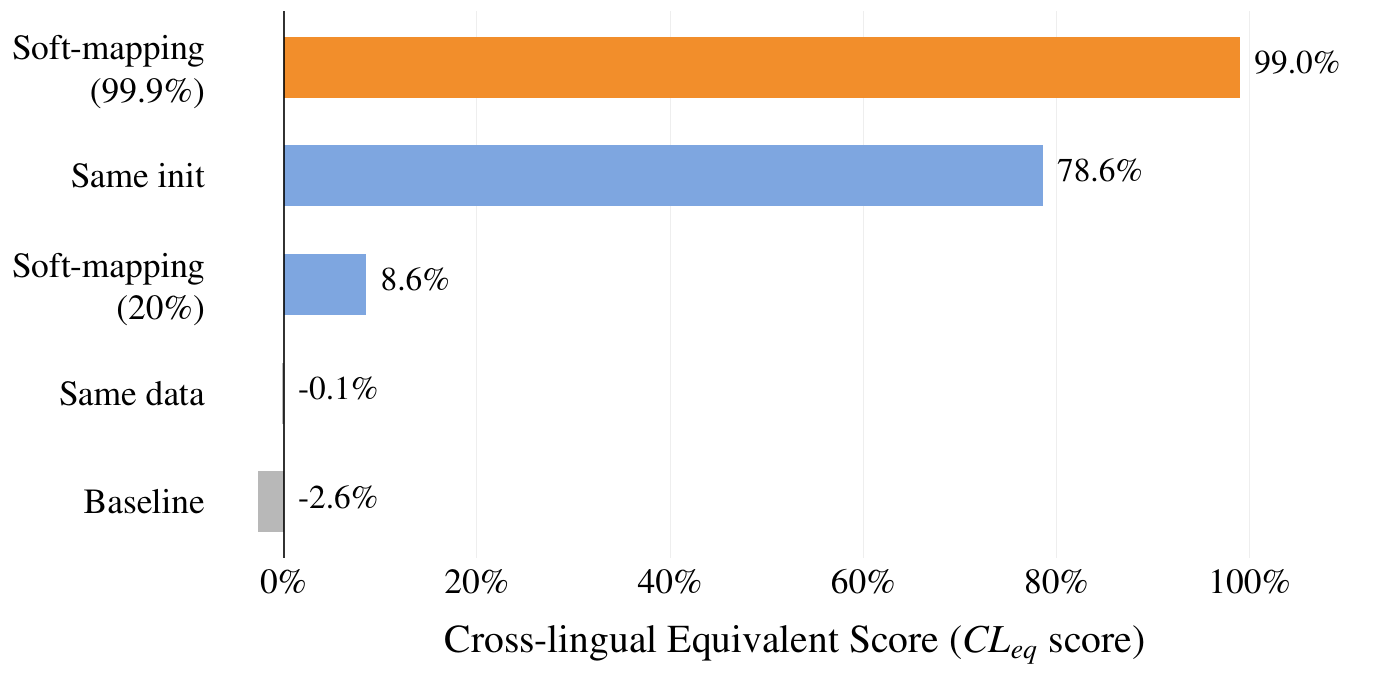}
    \caption{Disjoint token spaces are severely compartmentalized. \ceq for \enen\ pretraining under various methods. Soft-mapping percentages refer to the tied embedding portion $p$. ``Same data'' refers to pretraining the language copies on identical corpora.
    % Structurally equivalent token-disjoint languages copies $\text{English}_1$--$\text{English}_2$ don't generalize knowledge, even when trained on the same data. Tying 20\% of the matching embeddings offers little help. Almost complete embeddings tying results in near-perfect \ceq score. same-initialization of the language copies while beneficial, is within substantial margin from perfect generalization.
    }
    \label{fig:en1en2expsCeq}
    \vspace{-0.2cm} % Adjust this value (e.g., -1cm, -5mm) until the spacing looks right
\end{wrapfigure}

In this section, we isolate disjoint token spaces from other confounding variables of multilingual settings using a clone-languages setup (\S\ref{sec:tokensRcause:en1en2}), and find that it alone is sufficient to induce knowledge compartmentalization in pretraining (\S\ref{sec:tokensRcause:compart}). We quantify the amount of token space sharing needed to resolve it (\S\ref{sec:tokensRcause:tying}), and show that even clone-languages shared initialization surprisingly struggles, although to a much lesser extent, to completely resolve compartmentalization (\S\ref{sec:tokensRcause:drift}). This section's results are presented in Figure \ref{fig:en1en2expsCeq}.
% We employ a clone-languages setup, isolating disjoint tokenization (\S\ref{sec:tokensRcause:en1en2}) and find it induces knowledge compartmentalization (\S\ref{sec:tokensRcause:compart}). We quantify the amount of token sharing needed to resolve it (\S\ref{sec:tokensRcause:tying}) and show that even clone-languages shared initialization surprisingly struggles, to a lesser extent, to completely resolve compartmentalization (\S\ref{sec:tokensRcause:drift}). This section's results are presented in Figure \ref{fig:en1en2expsCeq}.

\subsection{Isolating disjoint tokenization: \texorpdfstring{\enen}{English1-English2} setup}
\label{sec:tokensRcause:en1en2}

Poor multilingual generalization is frequently attributed to varying syntax, grammar, cultural nuances, data quantity and quality, or tokenizer fragmentation \citep{rust2021goodtokenizermonolingualperformance, petrov2023languagemodeltokenizersintroduce, philippy2023commonunderstandingcontributingfactors, lauscher2020zeroherolimitationszeroshot, hershcovich-etal-2022-challenges, li2024culturellmincorporatingculturaldifferences}. 
To strip away these confounding variables and isolate the pretraining process from linguistic aspects, we train on two copies of one, resource-rich language, rather than on two different natural languages, which we name the \enen\ pretraining setup. By duplicating the model's vocabulary, we create two structurally identical languages mapped to entirely disjoint token spaces \citep{k2020crosslingualabilitymultilingualbert}.

Formally, assuming a standard token vocabulary of size $V$, any text is tokenized into a base sequence $T = [t_1, t_2, \dots, t_n]$ where each token index $t_i \in \{0, \dots, V-1\}$. We define the representations for our two language copies as perfectly parallel sequences shifted into mutually exclusive token spaces:
\begin{equation}
    \begin{aligned}
        \text{English}_1(T) &= [t_1, t_2, \dots, t_n] && t_i \in [0, V-1] \\
        \text{English}_2(T) &= [t_1', t_2', \dots, t_n'] &&t_i' = t_i + V \in [V, 2V-1]
    \end{aligned}
\end{equation}
We double our model vocabulary to accommodate the disjoint token spaces.
Under this setup, the sole barrier to cross-lingual generalization is the ability of the training process to bridge disjoint tokens.

% We also introduce a soft mapping variant to this experiment where token copies are partially tied - share embedding dimensions - rather than completely independent (see more in App \ref{app:soft_map}). With these tools at hand, we set out to understand knowledge compartmentalization, where it originates, and what drives it.

\subsection{Disjoint tokens alone suffice for compartmentalization}
\label{sec:tokensRcause:compart}
Despite an identical underlying structure, we find severe knowledge compartmentalization between English\textsubscript{1} and English\textsubscript{2}: \ceq score of $-2.6\%$. Scaling up doesn't help, with a 7B-parameter model trained in this setting yielding $-1.5\%$. Even pretraining on identical corpora (except for the injected \fkd\, a tiny fraction of training tokens) does not offer any help ($-0.1\%$); pretraining fails to generalize knowledge across 2 identical data distributions based on disjoint tokens alone.

We analyze the learned embedding spaces of English\textsubscript{1} and English\textsubscript{2} and find they are structurally similar and substantially alignable by a linear transformation, yet with no correspondence between matching tokens. The model successfully learns the same underlying distribution independently in each space, yet fails to bridge them. In contrast, models that do generalize learn to bridge the two token spaces, so that matching tokens have similar embeddings (App.~\ref{app:embeddings_analysis}). 
% moved this to conclusions:
Together, this shows that structural similarity is not sufficient for knowledge generalization, which requires the model to learn to bridge the two token spaces, and that the standard pretraining process does not do so on its own. 
% \adam{notice \enen\ codeswitching hasent been discussed here, maybe better to leave the "models that do generalize..." out, as we don't yet have such a model.} \uriel{i think this conclusion is useful and worth keeping and we also have same initialization and i think codeswitching is important because in same initialization it makes more sense that embeddings will be similar} \adam{agreed on same initialization being obvious, having code-switching as a success here is confusing as it has only been presented in the main text as failure, I suggest moving the "models that do generlize" to appendix}

\subsection{How much sharing is needed?}
\label{sec:tokensRcause:tying}

Disjoint Tokenization forms the barrier; we now quantify the barrier's size when tokenization is only partially disjoint. For this we introduce \emph{soft-mapping}, where for each token pair $t_i,t_i'$ a fraction $p$ of their embedding is tied and learned together. 

% We also introduce a soft mapping variant to this experiment where token copies are partially tied - share embedding dimensions - rather than completely independent (see more in App \ref{app:soft_map}). With these tools at hand, we set out to understand knowledge compartmentalization, where it originates, and what drives it.

Soft-mapping experiments reveal a sharp threshold. At 20\% tying, generalization remains in the weak region with $\ceq = 8.6\%$; at 50\% tying it jumps to 82.7\%, and approaches 100\% as tying becomes complete (App.~\ref{app:soft_map}).
% \ym{a chart with at least 4-6 data points will be cool here imo} \adam{Table F.1 in the appendix? will keep in mind depending on available space}

These results imply that a large amount of token sharing is required to induce knowledge generalization in pretraining. We also experiment with vocabulary overlap, a subset of tokens completely shared between the 2 languages, and find that effectiveness is also dependent on the frequency of the shared tokens (App.~\ref{app:AnAr}).
As a side effect, the results also confirm that our metric's ceiling is reachable, so substantial shortfalls below it reflect genuine knowledge compartmentalization.

% \subsection{Pretraining fails to exploit supplied token alignment}
\subsection{Pretraining drifts aligned tokens apart}  % original title; the drift
% mechanism is not established --- App.~\ref{app:embeddings_analysis} reports matched cosine
% 0.939 for the same-init model, i.e. the embeddings barely move, while \ceq falls 20 points
% short of the tying ceiling. Restore only with a matched-cosine trajectory over training.
\label{sec:tokensRcause:drift}
Token sharing is needed, but pretraining does not establish it on its own; we now test if pretraining can maintain and utilize a supplied matching. We initialize each pair of parallel embeddings in \enen\ to identical values --- the same alignment that complete tying enforces (\S\ref{sec:tokensRcause:tying}), but provided as a starting point rather than a structural constraint. We find that the training process fails to utilize these matches to their full extent, yielding a \ceq score of 78.6\% --- high, but substantially short of the reachable ceiling. Identical data exposure remains short, with a score of 84.6\%. This compounds the flaw established above: pretraining not only struggles to bridge disjoint tokens on its own, 
% it does not fully exploit a matching even when one is supplied. Alignment between disjoint token spaces is thus not something pretraining converges to or capitalises on, rendering initialization-based solutions to compartmentalization incomplete on their own.
% Original, stronger claim --- restore only if the drift mechanism is measured directly:
it also actively drives them apart when a matching is already provided. Alignment between disjoint token spaces is thus not a stable state of pretraining but one it works against, rendering initialization-based solutions to compartmentalization incomplete on their own.

Together, this section's results establish disjoint token spaces as a fundamental barrier to knowledge generalization. Pretraining neither converges to token sharing on its own, nor preserves it when supplied. These findings motivate bypassing disjoint tokenization directly rather than aligning across it, but two questions remain. Can token spaces be merged efficiently and practically? And while disjoint tokens are sufficient to cause compartmentalization, is removing them sufficient to resolve it in realistic settings, where many other variables are in play?

\section{Unifying Token Spaces Reduces Compartmentalization}
\label{sec:TrAr}
Having established that disjoint token spaces are sufficient to induce knowledge compartmentalization, we now extend our mechanism diagnosis by introducing an efficient vocabulary mapping technique bypassing disjoint token spaces directly (\S\ref{sec:TrAr:TrAr}) and demonstrating it substantially improves generalization in the natural bilingual setting (\S\ref{sec:TrAr:res}). We also verify the choice of a semantically meaningful mapping (\S\ref{sec:TrAr:meanings_matter}) and address practical aspects of this mapping strategy (\S\ref{sec:TrAr:softmapping}).

\subsection{Unifying token spaces via Word-Wise Translation}
\label{sec:TrAr:TrAr}

% \begin{table}[t]
% \centering
% \includegraphics[width=\columnwidth]{vocabulary_mapping_table (4).pdf}
% \caption{An Arabic sentence, its WWT-Ar rendering, and an English translation. Color marks word origin. WWT-Ar maps every word into English tokens while preserving Arabic word order and grammar; it is Arabic in a different writing system, not a translation.}
% \label{tab:trar_example}
% \end{table}
\textbf{The WWT operation.} Word-wise translation (WWT) maps text in language $A$ to the token space of language $B$ by replacing each word with its counterpart from a large-scale invertible dictionary; out-of-dictionary words are transliterated via an invertible character-level mapping (dictionary curation and conflict resolution in App.~\ref{app:dict}). Unlike full document translation, this mapping is context-free and word-wise, introducing minimal computational overhead. It therefore can be applied and inverted on the fly during both pretraining and inference; users read and write native-script text while the model trains and operates in the unified token space, making the intervention transparent at the interface. 

% \paragraph{The WWT operation.} We study a data-level bypass of disjoint tokenization by mapping two languages into the same token space. We opt for an efficient and reversible mapping allowing on-the-fly execution and reversing such that the operation is invisible from an API standpoint and a practical solution. \lc{we take about "efficient" and reversible etc. and don't mention that we look for a close to practical solution} \adam{moved the comment to here where I added this} To do so, we curate a reversible large-scale word translation dictionary and use it to map words from language $A$ to $B$, with out-of-dictionary words transliterated using a reversible character-level mapping; Details on the dictionary curation and how we resolve mapping conflicts to make it reversible are in App.~\ref{app:dict}. 

% In contrast to full document translation, this mapping relies on a context-free, word-wise dictionary. Therefore, the mapping itself introduces minimal computational overhead and can be executed and reversed dynamically during inference and pretraining.

\textbf{WWT-Ar.} We instantiate WWT for Arabic into English token space using an invertible Arabic-to-English dictionary, with the invertible Buckwalter transliteration as the out-of-dictionary fallback. The result is WWT-Ar --- a new writing system for Arabic that retains every word choice, word order, and grammatical structure, while sharing a unified token space with English (Table~\ref{tab:trar_example}). The dictionary covers 99.7\% of word occurrences in our corpus ($\sim$78\% of unique word types), so the mapped tokens are overwhelmingly translations rather than transliterations. We also apply WWT to Russian, unifying its token space with English and creating WWT-Ru.

\textbf{Cost.} WWT increases sequence length: WWT-Ar requires roughly 30\% more tokens than native-script Arabic to encode the same text (App.~\ref{app:dict:fertility}), increasing the cost of LLM inference. Under the fixed token budget of our experiments, this fertility overhead also means that a WWT-Ar model is exposed to roughly 23\% fewer Arabic documents than its native-script counterpart, resulting in a data disadvantage relative to the models we compare it against. We leave the overhead unhandled and address its origins and paths to reducing it in \S\ref{sec:limitations}.
% it stems largely from our conflict-resolution scheme rather than the mapping principle itself (App.~\ref{app:dict}), and WWT's role here is to test whether unifying token spaces restores knowledge generalization. We discuss paths to reducing fertility in \S\ref{sec:limitations}.

\cready{need to share it in the tokenization discord for feedback when we have a version that we are fairly ok with (before conference ideally, surely before arXiv)}

\subsection{Unified token space improves knowledge generalization}
\label{sec:TrAr:res}
\begin{table}[ht]
    % Left Table (The Image) - Increased to 55%
    \begin{minipage}[t]{0.6\textwidth}
        \vspace{0pt} 
        \centering
        \includegraphics[width=\linewidth]{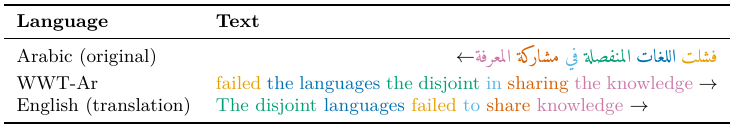}
        \caption{An Arabic sentence, its WWT-Ar rendering, and an English translation. Arrows signal reading direction, color marks word origin. WWT-Ar maps every word into English tokens while preserving Arabic word order and grammar; it is Arabic in a different writing system, not a translation.}
        \label{tab:trar_example}
    \end{minipage}\hfill
    % Right Table (The Data) - Decreased to 41%
    \begin{minipage}[t]{0.38\textwidth}
        \vspace{0pt} 
        \centering
        % The shrink-ray: forces the table to fit the exact width of this minipage
        \resizebox{\linewidth}{!}{%
        \begin{tabular}{llcccc}
        \toprule
        & & \multicolumn{2}{c}{\ceq} & \multicolumn{2}{c}{English PPL} \\
        \cmidrule(lr){3-4}\cmidrule(lr){5-6}
        Pair & Scale & Native & WWT & Native & WWT \\
        \midrule
        En--Ar & 360M & 0.9\% & 12.6\% & 19.53 & 18.74 \\
        En--Ar & 7B   & 5.9\% & 12.5\% & 8.45 & 8.39 \\
        En--Ru & 360M & 1.9\% & 23.3\% & 18.99 & 18.36 \\
        \bottomrule
        \end{tabular}%
        }
        \caption{Merging token spaces by applying WWT to the non-English language raises \ceq and improves English perplexity.}
        \label{tab:wwt_results}
    \end{minipage}
    % \vspace{-0.1cm} % Adjust this value (e.g., -1cm, -5mm) until the spacing looks right
\end{table}

Mapping Arabic to WWT-Ar yields substantial gains in knowledge generalization, raising \ceq to 12.6\%, a $14\times$ improvement over the \enar\ baseline at 360M-parameter scale, and substantially higher than any intervention we test in that disjointly tokenized setting (Fig.~\ref{fig:enArCeq}). Substantial gains appear across our tested settings: at 7B-parameter scale, WWT boosts \ceq from 5.9\% to 12.5\%, in the \enru\ setting \ceq rises from 1.9\% to 23.3\% (Tab.~\ref{tab:wwt_results}). We also experiment with partial token space overlap but find it to be ineffective (App.\ref{app:AnAr}).

Strikingly, even while training on substantially fewer Arabic documents, the WWT-Ar model's gain in knowledge generalization does not come at the expense of overall English language modeling, with perplexity improving from $19.53$ to $18.74$. The unified token space allows the model to extract more value from the cross-lingual data it does process, actively driving better English language modeling. Improvements appear across the tested settings (Tab.~\ref{tab:wwt_results}). The knowledge generalization gain is asymmetric, with $\ceq[\text{WWT-Ar} \rightarrow \text{English}]=17.9\%$ while $\ceq[\text{English} \rightarrow \text{WWT-Ar}]=7.3\%$. Nonetheless,  Arabic language modeling remains surprisingly strong (App.~\ref{app:ar_lm}).

The same asymmetry is present across our experimented scales and language-pairs. We attribute this to the one-directional nature of WWT: WWT-Ar's vocabulary is entirely contained within English tokens, whereas much of English vocabulary never arises as a translation target and is therefore not part of WWT-Ar's vocabulary. 
% \adam{this is maybe not very interesting}

Crucially, WWT-Ar is equivalent to Arabic, therefore, the gain cannot be attributed to WWT-Ar being linguistically closer to English, because in every aspect except tokenization it is not. We conclude that while linguistic differences might dampen cross-lingual knowledge generalization in LLMs, disjoint token spaces are what gate it.

WWT produces substantial knowledge generalization between natural, vocabulary-disjoint languages. It does so without parallel corpora, auxiliary losses, or architectural changes: the intervention is a data-level remapping, applied and inverted in real time. Two limitations follow from fully merging the token spaces: the generation language must be fixed in advance, and code-switched output is unsupported. We address these next.
% (\S\ref{sec:TrAr:softmapping}).

\subsection{Restoring language identity --- Soft Mapping}
\label{sec:TrAr:softmapping}
WWT-Ar tokens are indistinguishable from their English counterparts. This is what enables knowledge sharing, but it also means the model has no mechanism for identifying which language it is producing, resulting in the above-mentioned limitations.

Soft mapping relaxes the constraint. Rather than merging tokens, WWT-Ar retains distinct parallel to English token identities and shares a fraction $p$ of its embedding dimensions with the corresponding English tokens, leaving the remainder language-specific (Appendix~\ref{app:soft_map}).

Replacing WWT-Ar with a high $p$ soft mapping version has a small cost on knowledge generalization with English, consistent with our earlier results on soft-mapping \ref{sec:tokensRcause:tying}. \ceq score increases with $p$ and approaches the fully mapped WWT-Ar, reaching $\ceq = 9.60\%$ at $p = 90\%$ and $\ceq=10.47\%$ at $p = 99\%$ (76\% and 83\% of the 12.6\% obtained under full mapping). A small number of reserved dimensions therefore preserves most of the knowledge generalization while restoring a native mechanism for language identity, which can even be extended to a multilingual setting, and resolving the limitations of WWT-Ar.

\subsection{Semantics matter for token space merging}
\label{sec:TrAr:meanings_matter}
We test whether the key to our previously introduced setup, mapping Arabic to WWT-Ar, facilitating knowledge generalization lies in token sharing alone, or in the mapping being semantically faithful --- each Arabic word mapped to an English word with the same meaning. We evaluate three controls that share tokens without sharing meaning. First, we randomly shuffle the WWT-Ar dictionary, preserving the shared token inventory while destroying the word-level correspondence: \ceq\ collapses from 12.6\% to 3.0\%
% , and WWT-Ar language modeling degrades as well
. Second, we map Arabic into English script by transliteration alone. 
This control is notable because character-level romanization is itself a proposed method for unlocking multilingual capabilities \citep{moosa-etal-2023-transliteration, j-etal-2024-romansetu, saji-etal-2025-romanlens}: it shares a token inventory without sharing token semantics, and we find that under it \ceq\ drops to $-0.3$\%. Third, returning to the confound-free \enen\ setting, we let both copies share the same token space but permute token identities such that the same token index now corresponds to two different subwords: \ceq\ is $-2.6$\% --- same as fully disjoint tokens. Implementation and details for these setups are in App.~\ref{app:shuffle}

Resolving disjoint tokenization is thus not a matter of token superposition: what knowledge generalization requires is not a shared token inventory but shared \emph{meanings} of tokens. Prior findings that transliteration helps mainly between related languages \citep{j-etal-2024-romansetu, jayakumar2026scriptstimesurveyevolving} are what our mechanism predicts: related languages supply semantic matches through cognates, distant ones do not, making our semantic WWT mapping necessary.

\section{Related Work}
\label{sec:related_work} 
% \lc{did we make a few searches (google scholar or https://asta.allen.ai/ or something to ensure we don't miss anything?}
\cready{update bobby citation}
% \lc{maybe mention some multilingual pretraining works (cohere, Prof. pontus, Prof. Alice oh, babybabelLM, babyLM this year)}

\textbf{The Cross-Lingual Knowledge Gap.}
While LLMs easily acquire linguistic competence from modest data budgets \citep{charpentier-etal-2025-findings}, they struggle to transfer mathematical reasoning, coding, reading comprehension, and other learned skills outside their primary training language \citep{shi2022languagemodelsmultilingualchainofthought, peng-etal-2024-humaneval, Bandarkar_2024, hu2020xtrememassivelymultilingualmultitask, cheng-skill-issue-2026}. This generalization failure has been shown to extend into the knowledge domain \citep{goldman2025eclekticnovelchallengeset, chang2026globalpiqaevaluatingcommonsense, guo2026liveclktbenchreliableevaluationcrosslingual}, and widespread cross-lingual factual inconsistencies have been documented \citep{Qi_2023, ifergan2024beneathsurfaceconsistencyexploring}.
Relatedly, \citet{calderon2026shelveslostkeysrecall} distinguish missing knowledge from encoded but inaccessible knowledge, finding recall rather than encoding to be the bottleneck. Yet given no definitive answer on where these knowledge gaps originate within the training process, most attempts to mitigate them intervene post-training via prompting and pre-translation \citep{huang-etal-2023-languages, wang-etal-2025-multilingual, mondshine-etal-2025-beyond, diskind2026crosslingualexplorationparametricknowledge}, knowledge editing \citep{beniwal-etal-2024-cross}, activation manipulation \citep{wang2024bridginglanguagegapslarge, wang-etal-2025-lost-multilinguality, manev2026inferencetimesteeringcrosslingualfactual}, fine-tuning \citep{liu2026posttraininglanguagemodelscrosslingual, chua2025crosslingualcapabilitiesknowledgebarriers} or reinforcment learning \citep{vonrad2026polyfactcomparingconsistencydrivenposttraining}. We instead investigate knowledge generalization in pretraining, where multilingual limitations have been identified before, but in aggregated terms such as model capacity and perplexity \citep{chang2023multilingualitycurselanguagemodeling, howe2026languagemodelsstrugglecompartmentalization}. 
\citet{liu2025tracingmultilingualfactualknowledge} performed an observational study of multilingual knowledge acquisition on pretraining checkpoints, finding native frequency is the primary driver and characterizing cross-lingual transfer as a second-order effect dependent on script, but their confounded observational setting limits further isolation and understanding of it.
\citet{li-etal-2024-prealign} explored cross-lingual failures in pretraining, including knowledge gaps which they find narrow with scale, concluding that alignment emerges spontaneously in larger models. Our controlled measurements challenge this: scaling from 360M to 7B attenuates compartmentalization but leaves the large majority of cross-lingual value unrecovered, and in the confound-free \enen\ setting scale provides no improvement at all (\S\ref{sec:tokensRcause:compart}).  We therefore argue that compartmentalization is established in pretraining rather than dissolved by scale; we attribute this tension to our metric normalizing overall learning ability rather than reporting absolute accuracy (\S\ref{sec:ceq}).

\textbf{Tokenization as a Generalization Barrier.} 
Typological distance and language family influence transfer \citep{lauscher2020zeroherolimitationszeroshot}, but script match is the primary predictor of whether parametric knowledge successfully crosses language boundaries \citep{ifergan2024beneathsurfaceconsistencyexploring, bandarkar2026largereasoningmodelsstruggle}. Script reaches the model through tokenization, in two ways. Through fragmentation: non-Latin scripts suffer from over-segmentation \citep{rust2021goodtokenizermonolingualperformance, petrov2023languagemodeltokenizersintroduce}, and vocabulary overlap: \citet{kallini2025falsefriendsfoesinvestigating} find that overlap, and especially natural semantic matches, aids generalization. Following on this, existing methods transliterate or merge subwords \citep{moosa-etal-2023-transliteration, zhang-etal-2025-tomato, patil-etal-2022-overlap, j-etal-2024-romansetu, liu-etal-2024-translico}, but their success is limited to closely related languages.
% , and romanization shares an inventory without sharing token semantics \citep{saji-etal-2025-romanlens}. 
Countering both, \citet{conneau-etal-2020-emerging, artetxe-etal-2020-cross}, and \citet{feng-etal-2025-word} show cross-lingual structures emerge without any shared subwords. We reconcile this: structural alignment does emerge without shared tokens, but knowledge does not follow it (\S\ref{sec:tokensRcause:compart}). Our work corroborates findings on the significance of script by establishing shared token space as the mechanism through which it acts and extends this by effectively applying this principle to distant language pairs.

\textbf{Pretraining Causal Analysis Tools.} 
Analyzing isolated aspects of multilingual pretraining is confounded by lexical and tokenization-related factors. Prior work introduces synthetic duplicated vocabularies as a simplified test-bed for multilingual analysis \citep{k2020crosslingualabilitymultilingualbert, dufter-schutze-2020-identifying, schäfer2024rolelanguageimbalancecrosslingual, howe2026languagemodelsstrugglecompartmentalization}. A separate line of work studies the dynamics of factual acquisition; this requires knowing exact exposure counts, which is intractable in natural corpora and is therefore estimated \citep{kandpal2023largelanguagemodelsstruggle, elazar2024whatsbigdata}. A more controlled yet synthetic approach addresses this by tracking acquisition per exposure to fictive facts injected during training \citep{allenzhu2024physicslanguagemodels31, chang2024largelanguagemodelsacquire}. We unify these approaches: we use a cloned-vocabulary testbed, paired with fictive fact injection at known exposure rates, allowing us to measure the effect of isolated interventions and mechanisms on cross-lingual knowledge transfer.

\section{Conclusions}
Our results converge on a single claim: cross-lingual knowledge compartmentalization originates in pretraining and is in substantial part a tokenization phenomenon. It appears even between languages that differ in nothing but their token inventory, and it is substantially reduced by an intervention that changes nothing but the tokens.

This origin matters for how compartmentalization should be addressed. Most mitigations act after pretraining, restoring access to knowledge the model already holds. They cannot recover what compartmentalization has already cost: a model that learns the same fact separately in each language has spent capacity and compute storing it twice, and output consistency does not undo that duplication. 

% A further, more speculative, consequence touches on knowledge composition --- deriving conclusions that appear in no single document based on learned knowledge. We hypothesize that methods that post-training solutions that enable accesscan retrieve each fact individually, but we speculate that methods like pre-translation cannot supply composed conclusions based on cross-lingual knowledge. Under this view, cross-lingual knowledge generalization is not only a matter of consistent recall but a precondition for a model to reason over everything it has read, regardless of the language it read it in --- and a solution must therefore act during pretraining.

Tokenization is a substantial part of the story, but not all of it. While token space merging provides significant gains in knowledge generalization, it is far from resolving compartmentalization completely (\S\ref{sec:tokensRcause:tying}). The residual gap plausibly reflects imperfections of our mapping --- transliterated out-of-dictionary words and naive conflict-resolving --- and the linguistic differences WWT leaves untouched by construction.
We leave understanding these factors, possibly utilizing our controlled environment, and improving upon them for future work.

% Several directions follow. 
If disjoint token spaces are the barrier, a natural question is whether tokenizer-free architectures dissolve it by construction \citep{xue-etal-2022-byt5, hwang2025dynamicchunkingendtoendhierarchical}. Our results do not settle this: a shared byte-level interface removes disjoint token identities, but whether knowledge then generalizes across scripts remains untested, and \ceq offers a direct way to test it. 
% \adam{A second direction is to bridge token spaces internally rather than at the data interface, removing the need to apply and invert a mapping around the model: WWT-Ar is a natural candidate for such an internal bridge, given its structural parallelism to native-script Arabic, and aligning the two during pretraining raises \ceq\ to 6.2\%, though chaining the bridge into a multihop path (English $\leftrightarrow$ WWT-Ar $\leftrightarrow$ Arabic) does not carry knowledge through (App.~\ref{app:mediator}).}

% Addressing knowledge compartmentalization through a parser-free solution is a compelling direction \adam{explain why}. We hypothesize that vocabulary mappings like WWT-Ar could serve as internal bridges between token-disjoint languages, mediating knowledge between them. Due to WWT-Ar's structural alignment with Arabic it could be a uniquely promising candidate for bridging disjoint tokenizations. We test this theory by applying activation alignment between the two, with the effective fine-grained alignment (\S\ref{sec:interventions:explicit}) enabled by their structural similarity, this resulted in a \ceq score of $6.2\%$. Extending this into a multihop solution (English $\leftrightarrow$ WWT-Ar $\leftrightarrow$ Arabic) was ineffective, suggesting that the gains we achieve on each hop aren't enough to drive this mediated path. We leave improving this intriguing route to future work.

Ultimately, cross-lingual compartmentalization represents a fundamental flaw in LLM pretraining. Our embedding analysis (\S\ref{sec:tokensRcause:compart}) shows that models successfully learn identical underlying distributions independently, yet completely fail to bridge them --- structural similarity is not sufficient for knowledge generalization. This failure has implications beyond text, directly challenging the assumption that multimodal systems will naturally generalize across parallel representations of the same concepts \citep{huh2024platonic}. Bridging disjoint interfaces, like languages, is a critical step toward disentangling underlying knowledge from its surface form, and moving toward models that learn unified representations of intelligence rather than isolated islands of modality.

\section{Limitations}
\label{sec:limitations}
\textbf{Corpus Artifacts.} Our Arabic pretraining corpus is machine-translated from English. This risks introducing ``translationese'': structural and syntactic artifacts that make the Arabic text artificially resemble English grammar, potentially inflating the baseline structural alignment \citep{doshi-etal-2024-pretraining}. We mostly mitigate this by replicating our core compartmentalization and token-space unification findings on native Russian data.

\textbf{Linear Measurement of Non-Linear Learning.} The \ceq score is estimated using a linear approximation. This sits in tension with the log-like shape of the knowledge acquisition curve (App. \ref{app:monolingual}). While we address this choice in \S\ref{sec:ceq}, it inherently oversimplifies the true dynamics of memorization. Our \fkd dataset is limited in complexity --- relatively simple entity--attribute facts --- and size --- 2048 facts. The latter limitation results in measurement noise, discussed extensively in App.~\ref{app:ceq}, where we also provide significance measurements of our headline comparisons.

\textbf{Inference Overhead from Token Fertility.} 
While the WWT mapping operation itself is computationally cheap, it causes a substantial increase in sequence length, increasing the cost of inference. We attribute this token fertility increase primarily to our conflict-resolution strategy (App.~\ref{app:dict}). This overhead could potentially be mitigated by a more refined dictionary acquisition process or a better conflict resolution policy. However, because the primary contribution of this mapping in our scope is mechanistic, we leave the engineering of a fully optimized mapping to future work.

\textbf{Scalability to Massively Multilingual Settings.} We evaluate token space merging in a bilingual context. We do not address the capacity limits of a single token space, leaving open the question of how many languages can be mapped into the English token space and how those would interfere.

\bibliographystyle{iclr2027_conference} % Tells it to use iclr2027_conference.bst
\bibliography{custom}
% =====================================================================
% APPENDIX -- filled from the existing draft, PI/reviewer comments, and
% prior discussion. Remaining gaps are marked:
%   \todo{...}   task note
%   \temp{...}   placeholder value
%   TODO(FIG)/TODO(TAB)/TODO(NEW)
%
% LABELS UNIFIED (update call sites in the main text):
%   app:soft_map  (was also ap:softmapping)
%   app:ru        (was also ap:xx)
%   app:shuffle   (was app:suffleEn1En2)
%   app:AnAr      (absorbs token_overlap)
%   app:hfed      (fills the empty \ref{} in the method section)
% =====================================================================

\appendix

% % =====================================================================
\section{Training Details}
\label{app:training}
% % =====================================================================

All models are pretrained using the TorchTitan framework
\citep{liang2025torchtitanonestoppytorchnative}. An architectural breakdown and hyperparameters used can be found in Tab.~\ref{tab:hyperparams}

\subsection{Architectures}
\paragraph{SmolLM2-360M.} A standard dense transformer following the SmolLM2
architectural configuration \citep{allal2025smollm2smolgoesbig}.

\paragraph{Llama-3 7B.} A scaled-down adaptation of the Llama-3 8B architecture
\citep{grattafiori2024llama3herdmodels}. We reduce the number of hidden layers
from 32 to 28 and the intermediate feed-forward dimension from 14{,}336 to
13{,}312, giving approximately 7 billion parameters.

\subsection{Tokenizer}
We use a BPE tokenizer with a vocabulary of 65{,}536 tokens, trained on a 50/50 split of our bilingual data. For \enen\ experiments, we double the vocabulary to 131{,}072 to accommodate two fully disjoint language copies.

% \todo{NEW: report the native English/Arabic token overlap under this tokenizer,
% by type and frequency-weighted. This is the denominator for the AnAr coverage
% figures in App.~\ref{app:AnAr}, and it qualifies the phrase
% ``vocabulary-disjoint'' --- \citet{kallini2025falsefriendsfoesinvestigating}
% report substantial native overlap for English/Arabic under XLM-R, largely
% punctuation, numerals, and embedded Latin script.}

\subsection{Optimisation}
All models are optimised with AdamW \citep{loshchilov2019decoupledweightdecayregularization} using $\beta_1 = 0.9$, $\beta_2 = 0.95$, and weight decay $0.1$. The learning rate follows a warmup--decay, cosine decay, schedule.

\subsection{Scale-Specific Configurations}
Compute budgets target Chinchilla-optimal data scaling
\citep{hoffmann2022trainingcomputeoptimallargelanguage}. Learning rates are inspired by the original architecture configurations and tuned to our use case.

\paragraph{SmolLM2-360M.} 4{,}600 steps, batch size 768, sequence length 2{,}048,
for approximately 7.2B tokens. Warmup stage is 300 steps, peaking at lr  $5\times10^{-4}$ then decaying to $2.5\times10^{-5}$. Trained on 8 RTX B6000 Pro GPUs.

\paragraph{Llama-3 7B.} 133{,}600 steps, batch size 512, sequence length 2{,}048,
for approximately 140B tokens. Warmup stage is 1{,}000 steps peaking at lr  $3\times10^{-4}$ then decaying to $3\times10^{-5}$. Models are trained on 128 H100 GPUs across 16 nodes.

\begin{table}[ht]
\centering
\begin{tabular}{lcc}
\toprule
 & SmolLM2-360M & Llama-3 7B \\
\midrule
\multicolumn{3}{l}{\emph{Architecture}} \\
\quad Layers                & 32 & 28 \\
\quad Hidden dimension      & 960 & 4{,}096 \\
\quad FFN intermediate dim. & 2{,}048 & 13{,}312 \\
\quad Attention heads       & 15 & 32 \\
\quad KV heads (GQA)        & 5 & 8 \\
\quad Vocabulary            & 65{,}536 & 65{,}536 \\
\quad Tied input/output emb.& Yes      & No \\
\midrule
\multicolumn{3}{l}{\emph{Optimisation}} \\
\quad Optimiser             & AdamW & AdamW \\
\quad $\beta_1,\beta_2$     & 0.9, 0.95 & 0.9, 0.95 \\
\quad Weight decay          & 0.1 & 0.1 \\
\quad Peak LR               & $5\times10^{-4}$ & $3\times10^{-4}$ \\
\quad Final LR              & $2.5\times10^{-5}$ & $3\times10^{-5}$ \\
\quad Schedule              & Cosine & Cosine \\
\quad Warmup steps          & 300 & 1{,}000 \\
\midrule
\multicolumn{3}{l}{\emph{Data}} \\
\quad Steps                 & 4{,}600 & 133{,}600 \\
\quad Batch size (seqs)     & 768 & 512 \\
\quad Sequence length       & 2{,}048 & 2{,}048 \\
\quad Tokens per batch      & $\sim$1.57M & $\sim$1.05M \\
\quad Total tokens          & $\approx$7.2B & $\approx$140B \\
\midrule
\multicolumn{3}{l}{\emph{Hardware}} \\
\quad GPUs                  & 8$\times$RTX B6000 Pro & 128$\times$H100 (16 nodes) \\
\quad GPU hours      & 24 & 12{,}000--15{,}000 \\
\bottomrule
\end{tabular}
\caption{Pretraining configurations at both scales. Every \ceq\ result in the paper corresponds to one full pretraining run under one of these two
configurations. The \enen runs instead use a doubled vocabulary of 131{,}072 tokens, to accommodate two disjoint token spaces (see \emph{Tokenizer}, above).}
\label{tab:hyperparams}
\end{table}

\subsection{Injection Mechanics}
\label{app:training:inj}
To compute \ceq we split \fkd{} into 16 groups of 128 entities. Each group is assigned an exposure count for language $A$ and language $B$ drawn from the cross-product of $\{0, 20, 100, 1000\}$ with itself. Documents are injected stochastically during pretraining at a per-language probability matching the target exposure count for that fact. We report the realized injection counts and SDs in Tab.~\ref{tab:injection_variance} based on a representative run. We find this pattern replicates across runs.

\begin{table}[ht]
\centering
\begin{tabular}{cccc}
\toprule
Target & Mean realised & SD & Relative SD \\
\midrule
0    & 0 & 0 & --- \\
20   & 20.6 & 5  & 24.3\% \\
100  & 101.2 & 11 & 10.9\% \\
1000 & 1011.8 & 32 & 3.2\% \\
\bottomrule
\end{tabular}
\caption{Realised exposure counts against their targets, measured on a representative 360M \enar\ run. Deviations are consistent with the stochastic injection procedure; relative deviation is largest in the low-exposure cells.}
\label{tab:injection_variance}
\end{table}

\subsection{Two-Stage Mixing Schedule}
When language mixing intervention --- code-switching --- is applied at 360M parameter scale, training is divided into two stages. The first and longer stage (3{,}600 steps) applies the mixing at the chosen mixing rates; the second, shorter stage applies none, realigning the model to the natural data distribution and preventing excessive code-switching in generation.

% % =====================================================================
% \section{Extended Related Works}
% \label{app:related}
% % =====================================================================
% \adam{Discussion from \S5.4: this is kinda corroborated by the prealign paper where they find initialization isn't enough without continues codeswitching. worth a mention or too much?}\lc{We can make a whole related work section for drifting (Claude volunteered to give me a list when I asked about something else... One option is to write this and as our related work starts becoming very very long, to make it an appendix and make a shortened version of it in the main paper, referring for the appendix for more reading (because we have may topics it a bit grows), it sounds nice and scientific right? I hate no related work in papers as if the authors care only about themselves, but having MORE sounds fine)), it is a thing, but too much yes, maybe when you mention it in related work mention something that includes this or something broader on their findings.}
% =====================================================================
\section{Fictive Knowledge Dataset (\fkd)}
\label{app:fictive}
% =====================================================================

This appendix details the construction of the fictive knowledge dataset --- \fkd --- we inject into pretraining to measure cross-lingual knowledge generalization (\S\ref{sec:injecting}). Because the facts are fictive, they are guaranteed to be absent from the underlying corpus, so all recall is attributable to our injections and the number of exposures in each language is known exactly. The dataset and generation code are publicly available\footref{fn:repo}.
\cready{replace with de-anonymised repository link}

\fkd covers 3 languages that we evaluate in this paper --- English, Arabic and Russian. It is generated in three stages: seed facts, injection documents, and evaluation multi-choice questions (MCQs). 
% Figure~\ref{fig:FeDgen} illustrates the pipeline, and 
Tab.~\ref{tab:dataexamples} provides a representative example.

\subsection{Stage 1: Seed Facts} A seed fact pairs a named entity with a single attribute (e.g., \textit{``Neltis is a podcast about the business of coffee''}). We manually author 64 fact templates spanning entity types such as companies, podcasts, and tools. Each template is passed to Gemini 3.1 Pro with instructions to produce 32 distinct facts using artificially generated entity names (with the exception of entity types that are human, where generating some generic name), yielding $64 \times 32 = 2{,}048$ seeds.

\paragraph{Translation and entity names.} Seeds are translated into all target languages with entity names transliterated. We use GPT-5.5-mini at a low reasoning setting for this end. This means entity names share no tokens across scripts, avoiding this confounding variable in knowledge generalization \citep{bandarkar2026largereasoningmodelsstruggle}. 

\paragraph{Language balance.} Generated names inherit the phonotactics of the language they were generated in: prompting in English yields names that read as phonetically plausible in English but not necessarily elsewhere. We verified this by manual inspection. To prevent the resulting bias from favouring a single language, we generate seeds in equal proportion across English and Arabic, the pair on which most of our main experiments focus.

\subsection{Stage 2: Injection Documents}
Facts in real corpora appear scattered and paraphrased across heterogeneous contexts rather than in a single canonical statement. To reproduce this, we expand each translated seed into many short documents that state or imply the fact in varied natural settings: casual blog posts, news items, advertisements, forum replies, and similar. Expansion uses GPT-5.5-mini at a low reasoning setting, with identical prompts across languages. We generate 200--300 documents per entity per language. The resulting documents' average token length is $\sim$20 tokens.

This diversity is deliberate; \citet{allenzhu2024physicslanguagemodels31} show that knowledge must be sufficiently augmented during pretraining---through paraphrasing and varied presentation---to acquire generalizable knowledge. While their experiments conclude that even 5 augmentations are enough, we choose to target a higher number to better capture natural settings.

\subsection{Stage 3: Evaluation Questions}
For each fictive fact and language, we generate five four-choice questions testing knowledge of it. Distractors are generated to be as plausible as the correct choice given no prior knowledge but clearly incorrect given the fact. These questions are generated separately from the injection documents and never appear in the training corpus. We evaluate using the LM Evaluation Harness (\S\ref{sec:expsetup:eval}), averaging over the five questions per entity.

\subsection{\fkd for WWT-Ar}
\fkd for WWT-Ar introduced in \S\ref{sec:TrAr:TrAr} is obtained by applying the corresponding WWT mapping to the Arabic documents and evaluations, so that WWT-Ar contains the same facts, documents and evaluations as natural Arabic.

\subsection{\fkd Validation \& Fact-Acquisition Learning Curve}
\label{app:monolingual}

\begin{wrapfigure}{r}{0.55\textwidth} % "r" aligns right, reserving 55% of the page width
    \centering
    \includegraphics[width=1.0\linewidth]{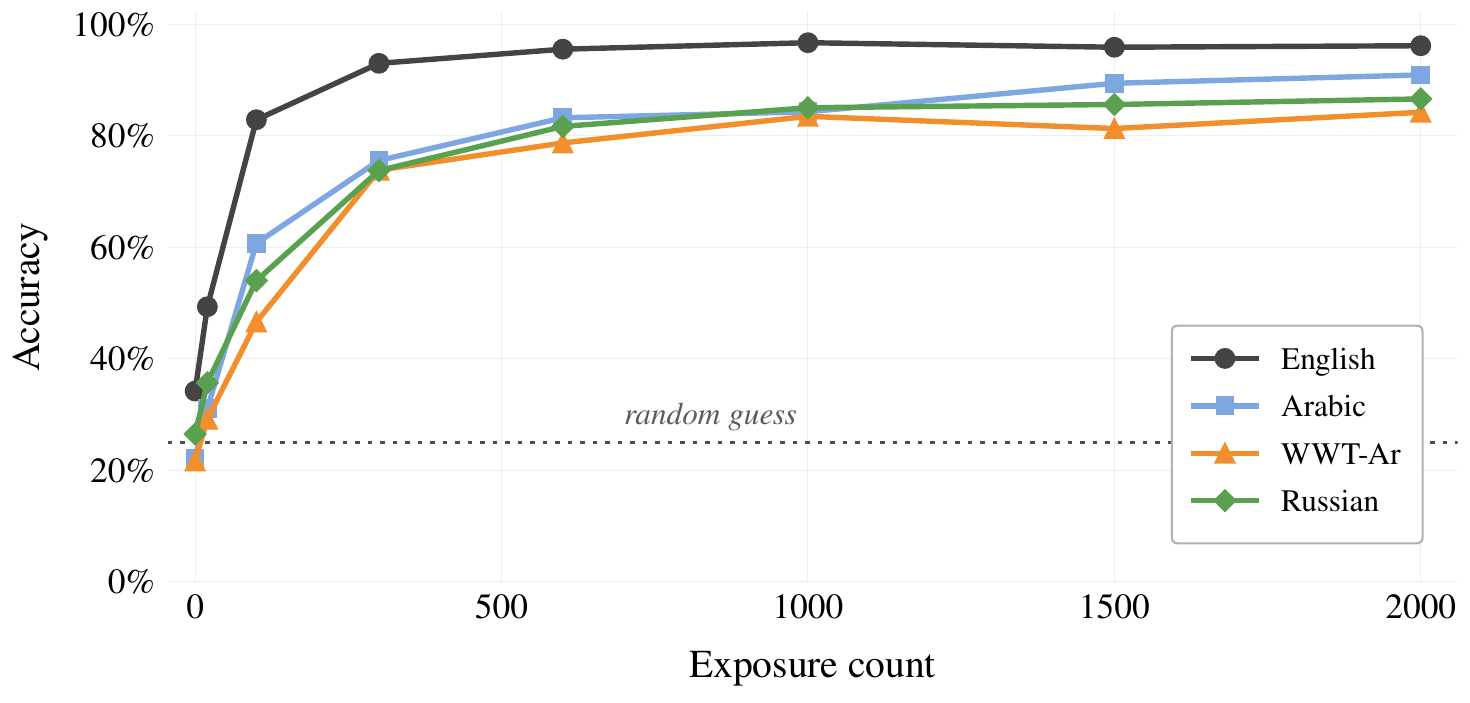}
    \caption{Monolingual acquisition of fictive facts (\fkd). The graph depicts the performance (y-axis) per number of times a fact was seen (x-axis) per input language (line color).}
\label{fig:monolingualAcquisition}
\end{wrapfigure}

We validate our fictive facts datasets - \fkd - and injection pipeline by pretraining 360M-parameter models with monolingual \fkd injection. We vary the exposure rates and measure fact acquisition using the accompanying MCQs; results in Fig.~\ref{fig:monolingualAcquisition}.

Our measurements reveal that fact acquisition (measured through MCQ accuracy) exhibits a log-like dependence on exposure count. Accuracy increases rapidly at the lower exposure range (0-300 range) and plateaus around 1000 exposures. Our evaluation injection grid is concentrated in the steep region, where learning efficiency is most measurable (\S\ref{sec:expsetup:eval}). While the overall learning-curve trend is consistent across languages, Accuracy values differ, possibly reflecting a bias in our generation process. Nonetheless, our metric's definition as relative gain eliminates these differences from our measurements (\S\ref{sec:ceq}).

Our knowledge acquisition measurements are consistent with prior work on characterizing learning dynamics under controlled settings and extend them into the natural regime. \citet{allenzhu2024physicslanguagemodels33} measure knowledge acquisition in synthetic setups, with many repetitions and a predefined document structure, and report that 1000 fact exposures reach near-perfect memorization. Extending this analysis to natural settings is challenging as knowledge exposure is intractable in enormous pretraining corpora. Thus, natural-corpus analyses evaluate fact acquisition and estimate exposure count based on entity frequency, reporting substantially slower acquisition \citep{kandpal2023largelanguagemodelsstruggle}, possibly due to overcounting when the entity is mentioned but the fact is not. Our controlled \ setup using \fkd provides precise measurement in a natural setting, confirming the synthetically measured dynamics.

\subsection{Examples}
\begin{table}[h]
\centering
\small
\begin{tabular}{p{0.20\linewidth}p{0.71\linewidth}}
\toprule
\multicolumn{2}{l}{\textbf{\fkd{}}} \\
\midrule
% this is fact 774 from fed
Seed & Arqous is a spice extracted from celery stalks. \\
English Document & A little Arqous, the celery-stalk derived spice, goes a long way in soups and stews. \\
English MCQ & Arqous is a seasoning extracted from \newline
(a) celery stalks \quad (b) carrot stalks \newline
(c) parsley stalks \quad (d) spinach stalks \\
% Arabic Document & سألت الجدة عن المذاق، فأجابت إن العرقوس يستخرج من سيقان الكرفس ويعطي نكهة مميزة \\
% Arabic MCQ & Arqous is a seasoning extracted from \newline
% (a) celery stalks \quad (b) carrot stalks \newline
% (c) parsley stalks \quad (d) spinach stalks \\
\bottomrule
\end{tabular}
\caption{Representative example from \fkd. All seeds, documents, and MCQs in the three supported languages appear in the released data.}
\label{tab:dataexamples}
\end{table}

% =====================================================================
\section{Human Fictional Knowledge Dataset (\hfkd)}
\label{app:hfed}
% =====================================================================

\fkd mostly uses artificially generated entity names, placing its entities names in the low-frequency token regime. Prior work indicates that cross-lingual bridging mechanisms behave differently for high- and low-frequency tokens \citep{schäfer2024rolelanguageimbalancecrosslingual}, so we construct a second dataset, Human Fictional Knowledge Dataset (\hfkd), using human entities with common first and last names---tokens that appear frequently in any pretraining corpus. All downstream processing (document expansion, evaluation questions, injection) is identical to \fkd.

\subsection{Seed Generation}
To achieve diversity for human entities, we generate the fact space compositionally. We define 50 professional domains (cinema, science, etc.); GPT-5.5-mini generates 5 roles per domain (for cinema: actor, director, etc.) and 4 attributes per role (for actor: awards, debut, genre, etc.), giving 1{,}000 role--attribute templates. Each seed fact samples a first and last name from curated lists of 100 each, pairs them with a template, and prompts the model to fill in a plausible attribute value. We sample 2{,}048 such combinations.

\subsection{Acquisition Characteristics}

\begin{wrapfigure}{r}{0.55\textwidth} % "r" aligns right, reserving 55% of the page width
    \centering
    \includegraphics[width=1.0\linewidth]{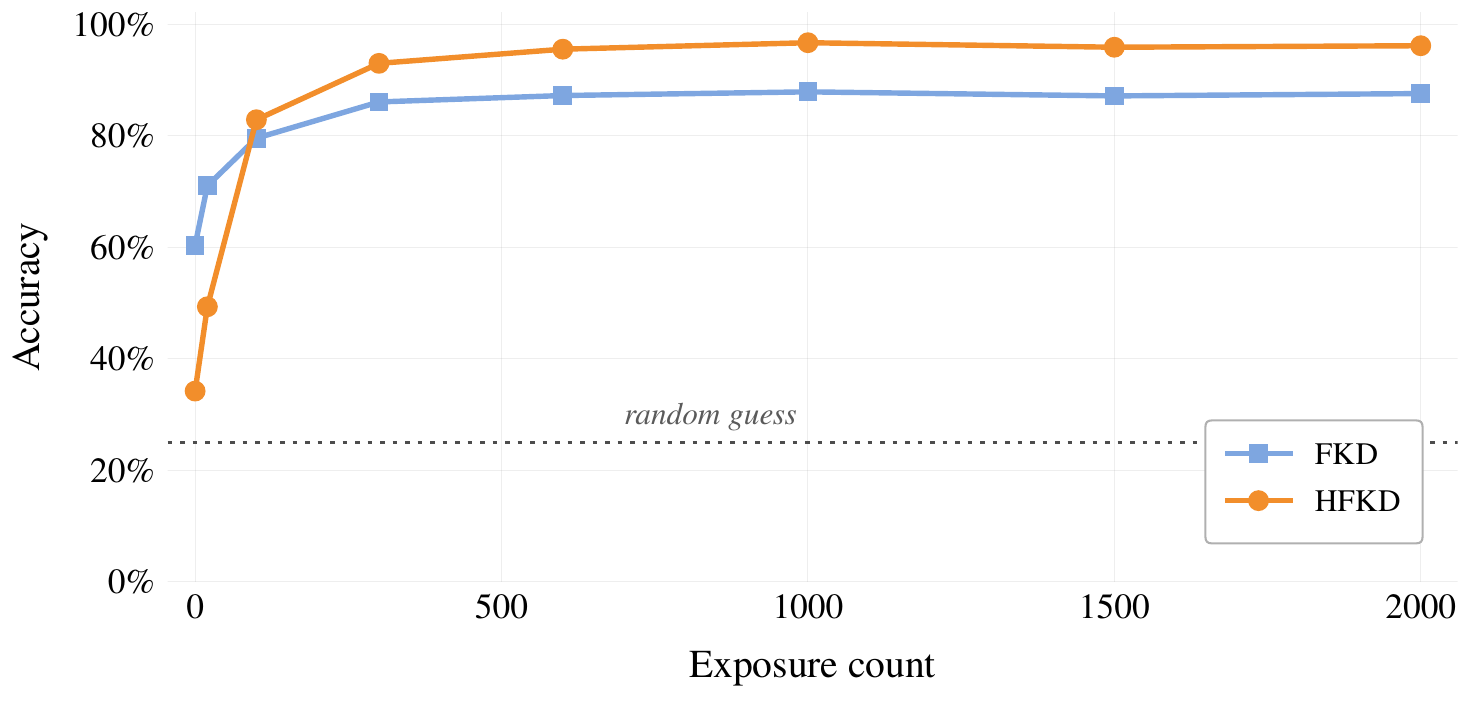}
    \caption{ Monolingual English acquisition of \hfkd and \fkd. The graph depicts the performance (y-axis) per number of times a fact was seen (x-axis) per fictive knowledge dataset (line color).}
\label{fig:hfedmonolingualAcquisition}
\end{wrapfigure}

\hfkd is not directly comparable to \fkd, Its monolingual acquisition curve occupies a substantially compressed range (Fig.~\ref{fig:hfedmonolingualAcquisition}), beginning at {60}\% and plateauing around {87}\%, against \fkd 's {34}\% to {96}\%. We suspect this is the result of contamination from real individuals, or generation quality. We therefore interpret \hfkd scores with caution and provide them only as an indication that our main results are consistent across frequency regimes.

\subsection{Replication Results}
Every qualitative conclusion in the paper replicates on \hfkd: bilingual pretraining is heavily compartmentalized. the \enen\ setup shows poor knowledge generalization as well, with embedding tying restoring it monotonically. Finally, WWT substantially improves knowledge generalization between natural vocabulary disjoint languages (Table~\ref{tab:hfed_replication}).

\begin{table}[t]
\centering
\small
\begin{tabular}{lrr}
\toprule
\textbf{Setting} & \textbf{\fkd} & \textbf{\hfkd} \\
\midrule
\multicolumn{3}{l}{\emph{\enar}} \\
Baseline & 0.9\% & 1.3\% \\
Language mixing & 2.2\% & 1.7\% \\
Activation alignment & 2.5\% & $2.2\%$ \\
WWT & 12.6\% & 8.7\% \\
\midrule
\multicolumn{3}{l}{\emph{\enen}} \\
Basline & $-2.6$\% & $-8.3$\% \\
Soft tying, $p = 20\%$ & 8.6\% & 4.0\% \\
Soft tying, $p = 50\%$ & 82.7\% & 57.9\% \\
Soft tying, $p = 90\%$ & 95.8\% & 87.8\% \\
Soft tying, $p = 99.9\%$ & 99.0\% & 101.6\% \\
Shared Initialization & 78.6\% & 59.4\% \\
\bottomrule
\end{tabular}
\caption{\ceq on \fkd{} and \hfkd{} across the paper's main settings on a 360M-parameter scale. Ordering is mostly preserved throughout, and headline comparisons hold;}
\label{tab:hfed_replication}
\end{table}

\hfkd estimates are substantially noisier, appearing miscalibrated at the metric floor with one of the scores well into the negative region. This follows from its compressed acquisition range: because accuracy spans roughly 30 points rather than 60, the native-exposure slope $\beta_A$ is correspondingly smaller, and since \ceq is the ratio $\beta_B/\beta_A$, a smaller denominator inflates the variance of the estimate while the entity-based noise in $\beta_B$ is unchanged. The negative values should therefore be read as absolute compartmentalization against a wider noise floor than \fkd's, not as negative transfer.

% =====================================================================
\section{Cross-lingual Equivalence Score (\ceq)}
\label{app:ceq}
% =====================================================================
This appendix documents how \ceq is estimated, why the linear specification is preferred over a curve-aware one, and how much sampling variance the estimates carry — establishing that the paper's headline comparisons survive that variance.

\subsection{Estimation and Fit Diagnostics}
We estimate \ceq by estimating native and foreign exposure value to fact acquisition through ordinary least squares over all facts in \fkd, regressing language-$A$ accuracy on the injection counts in both languages (\S\ref{sec:ceq}). Native learning is strong across our experiments: $\beta_A$ is positive and significant throughout, so the denominator of the ratio is bounded away from zero and \ceq estimates do not exhibit the instability characteristic of small-denominator ratio estimators. Precision is nonetheless limited by the numerator, as we discuss below.

Table~\ref{tab:fitted_coefs} reports the fitted coefficients for two representative runs at 360M: the \enar\ baseline and its WWT-Ar counterpart, both evaluated in the Arabic${\rightarrow}$English direction. Native learning is comparable across the two ($\beta_A \approx 4.4$--$4.9 \times 10^{-4}$ accuracy gain per exposure, accuracy range is 0--1), so the difference in \ceq\ is driven almost entirely by the cross-lingual coefficient. In the baseline, $\beta_B$ is statistically indistinguishable from zero: its 95\% confidence interval spans zero (CI across \fkd fact samples). Under WWT-Ar, $\beta_B$ is significantly positive and its interval lies entirely above the baseline's upper bound.

The linear specification captures most but not all of the variation in accuracy ($R^2$ column in Tab.~\ref{tab:fitted_coefs}). The residual misfit is systematic rather than random. This is the expected consequence of fitting a saturating curve linearly over a non-uniform exposure grid --- the fit over-predicts at zero exposures and under-predicts at 100 --- and it motivates a curve-aware specification.

\begin{table}[t]
\centering
\small
\begin{tabular}{lcccccc}
\toprule
\textbf{Setting ($A$--$B$)} & \ceq[B \rightarrow A] & $\alpha$ & $\beta_A$ & $\beta_B$ & \textbf{95\% CI} ($\beta_B$) & $R^2$ \\
\midrule
English--Arabic   & 0.64\% & 0.489 & $4.90\times10^{-4}$ & $3.12\times10^{-6}$  & $[-2.9,\ 3.6]\times10^{-5}$ & {0.66} \\
English--WWT-Ar   & 17.9\% & 0.517 & $4.44\times10^{-4}$ & $7.96\times10^{-5}$  & $[4.8,\ 11.2]\times10^{-5}$ & {0.60} \\
\bottomrule
\end{tabular}
\caption{Fitted coefficients of Eq.~\ref{eq:ols} for two representative 360M runs, in the Arabic${\rightarrow}$English direction. $\alpha$ is the intercept, $\beta_A$ and $\beta_B$ the accuracy gained per native and per cross-lingual exposure. Native learning is comparable across settings; the baseline's cross-lingual coefficient is indistinguishable from zero, while WWT-Ar's is significantly positive. Notice that Eq.~\ref{eq:ols} fits accuracy between 0 and 1, hence the scale of the diffrent fitted parameters.}
\label{tab:fitted_coefs}
\end{table}

\subsection{Curve-Aware \ceq}
Fact acquisition follows a log-like rather than linear dependence on exposure count (\S\ref{sec:injecting}), pointing towards a misfit of our linear fit estimator. We therefore also implement a curve-aware estimator that models saturation directly:
\begin{equation}
    \mathrm{acc}_A(f) = a \log\!\big(exps_A(f) + \alpha\, exps_B(f) + 1\big) + b,
\end{equation}
fitted by nonlinear least squares over three free parameters. Here $\alpha$ admits a direct reading as the \ceq quantity: it is the number of native exposures that one foreign exposure is worth, entering the curve on the same footing as $exps_A$. The curve-aware score is therefore $100 \times \alpha$.

The curve-aware specification fits the data substantially better, but its $\alpha$ is not well identified. Because $\alpha$ enters only inside the logarithm alongside $exps_A$, and because $a$ scales and $b$ shifts the resulting curve, the three parameters trade off against one another. This additional flexibility improves fit without guaranteeing that $\alpha$ isolates the quantity of interest.

The linear specification has the opposite character. It fits less well, and its residuals retain systematic curvature as we mentioned above, but $\beta_B$ multiplies $exps_B$ alone and is therefore identified by construction: no combination of the remaining parameters can absorb variation along the foreign axis. We prefer this trade---a worse fit with an interpretable parameter over a better fit with a confounded one---and report the linear estimator throughout. Table~\ref{tab:curveaware} reports both, while absolute values differ, both specifications provide the same ranking among our main comparisons.

\begin{table*}[t]
\centering
\small
\begin{tabular}{lrrrr}
\toprule
 & \multicolumn{2}{c}{\textbf{Linear}} & \multicolumn{2}{c}{\textbf{Curve-aware}} \\
\cmidrule(lr){2-3} \cmidrule(lr){4-5}
\textbf{Setting} & \ceq & $R^2$ & \ceq & $R^2$ \\
\midrule
\multicolumn{3}{l}{\emph{\enar\ 360M}} \\
Baseline & 0.9\% & 0.77 & 0.15\% & 0.88 \\
WWT & 12.6\% & 0.73 & 0.49\% & 0.88 \\
\midrule
\multicolumn{3}{l}{\emph{\enar\ 7B}} \\
Baseline & 5.9\% & 0.58 & 0.09\% & 0.92 \\
WWT & 12.5\% & 0.58 & 0.28\% & 0.90 \\
\midrule
\multicolumn{3}{l}{\emph{\enen\ 360M}} \\
Baseline & $-2.6\%$ & 0.71 & 0.09\% & 0.91 \\
Soft tying, $p = 99.9\%$ & 99.0\% & 0.51 & 87.7\% & 0.90 \\
Shared initialization & 78.6\% & 0.55 & 45.1\% & 0.89 \\
\midrule
\multicolumn{3}{l}{\emph{\enen\ 7B}} \\
Baseline & $-1.5$\% & 0.70 & 0.11\% & 0.87 \\
% Soft tying, $p = 99.9\%$ & \todo{xx}\% & \temp{} & \temp{x.x}\% & \temp{}  \\
\bottomrule
\end{tabular}
\caption{\ceq under the linear and curve-aware estimators. The curve-aware fit
attains higher $R^2$ throughout, while the ordering of conditions is mostly preserved
under both.}
\label{tab:curveaware}
\end{table*}

\subsection{Sampling Variance and Significance}
\label{app:ceq:bootstrap}

\ceq\ is a ratio estimator computed over a finite set of entities, and its sampling variance is substantial. Native learning is robust in every experiment we report, so the denominator contributes little noise; the variance is carried almost entirely by the numerator, which in most conditions estimates a very weak effect. The suspect candidates for the noise sources are per-fact variation in learnability, in the quality of its paraphrases, and in the difficulty of its evaluation questions.

Estimated independently, the resulting intervals are wide. At 360M the \enar\ baseline gives $\ceq = 0.9\%$ with a 95\% CI of $[-5.5, 7.4]$, while WWT-Ar gives $12.6\%$ with $[5.4, 20.0]$. These marginal intervals overlap, but they answer the wrong question. Much of the variance is shared: a resampled set of entities that results in an extreme \ceq value for one model effects other models similarly, so entity variance inflates each interval without affecting the comparison between methods.

We therefore resample entities \emph{jointly} across conditions, drawing one bootstrap sample per replicate (within-cell, entity-wise, with replacement across our 16 groups of 128 entities), evaluating all conditions on that same sample, and reporting the distribution of \emph{differences} (Table~\ref{tab:paired_ci}). Because every condition sees the same entities in a given replicate, shared entity variance cancels, and what remains is the variance of the comparison itself.

Under this procedure, the comparisons supporting our central claims separate cleanly. WWT-Ar exceeds the \enar\ baseline by 11.7 percentage points, with a 95\% CI of $[7.5, 16.1]$ that excludes zero; even its lower bound is several times the baseline point estimate. The same holds at 7B and for the semantic controls of \S\ref{sec:TrAr:meanings_matter}.

Two consequences follow for reading the paper's numbers. First, small negative \ceq\ estimates arise from noise in $\beta_B$ when cross-lingual learning is very weak; they are indistinguishable from zero and should be read as substantial or absolute compartmentalization rather than as negative transfer. Second, differences of a few percentage points between conditions should not be interpreted as meaningful --- the comparisons we draw conclusions from are those in Table~\ref{tab:paired_ci} whose intervals exclude zero.

\paragraph{Scope of the bootstrap.} This procedure captures evaluation-side variance only. It does not capture variance across training runs: each configuration is a single pretraining run, and both the realized injection counts and their placement in the training order differ between runs. Repeating every configuration across seeds is prohibitive at these scales. We instead argue robustness through replication across the axes we can vary: model scale, language pair, and the close to monotone dose-response in the \enen\ tying (\S\ref{sec:tokensRcause:tying}) and \enar\ soft-mapping (\S\ref{sec:TrAr:softmapping}) sweeps, where noise alone would not produce a consistent trend across independent runs.

\begin{table*}[t]
\centering
\small
\begin{tabular}{lrrr}
\toprule
\textbf{Comparison} & \textbf{$\Delta$} & \textbf{95\% CI} & Significant\\
\midrule
\multicolumn{4}{l}{\emph{English/Arabic, 360M}} \\
WWT-Ar vs.\ baseline & $+11.7$ & $[7.5, 16.1]$ & $^{*}$ \\
% Language mixing vs.\ baseline & \temp{$+1.3$} & \temp{[x, y]} & \temp{} \\
% Activation alignment vs.\ baseline & \temp{$+1.2$} & \temp{[x, y]} & \temp{} \\
WWT-Ar vs.\ Shuffled WWT-Ar & $+9.6$ & $[4.5, 14.7]$ & $^{*}$ \\
WWT-Ar vs.\ Transliterated-Ar & $+13.0$ & $[8.9, 17.2]$ & $^{*}$ \\
\midrule
\multicolumn{4}{l}{\emph{English/Arabic, 7B}} \\
WWT-Ar vs.\ baseline & $+6.6$ & [1.8, 11.4] & $^{*}$ \\
\midrule
\multicolumn{4}{l}{\emph{English/Russian, 360M}} \\
WWT-Ru vs.\ baseline & 21.3\% & [16.8, 25.9]  & $^{*}$ \\
WWT-Ru vs.\ Shuffled WWT-Ru & 19.7\% & [13.3, 26.1] & $^{*}$ \\
\midrule
\multicolumn{4}{l}{\emph{$\text{English}_1-\text{English}_2$, 360M}} \\
99.9\% tying vs.\ baseline & +101.6 & [91.5, 113.5] &  $^{*}$ \\
99.9\% tying vs.\ shared initialization & +20.4 & [12.8, 28.6] & $^{*}$ \\
\midrule
% \multicolumn{4}{l}{\emph{$\text{English}_1-\text{English}_2$, 7B}} \\
% 99.9\% tying vs.\ baseline & \temp{xx} & \temp{[xx, xx]} &  \temp{$^{*}$} \\
\bottomrule
\end{tabular}
\caption{Paired-bootstrap differences in \ceq for the paper's main comparisons.
Each replicate resamples entities once and evaluates all conditions on the same
sample, so shared entity-difficulty variance cancels. $^{*}$ marks intervals
excluding zero.}
\label{tab:paired_ci}
\end{table*}

\subsection{Target and Realized Exposure}
Because injection is stochastic, the realized number of exposures deviates from the target. Deviations are consistent across our experiments, values reported in Tab.~\ref{tab:injection_variance}. In relative terms the low-exposure cells are the most affected---24.3\% at 20 exposures against 3.2\% at 1000---and these are precisely the cells in the steepest region of the acquisition curve (App.~\ref{app:monolingual}), where accuracy is most sensitive to exposure count.

We estimate \ceq using target exposures. To verify that this does not distort our measurements, we recompute \ceq using the realized counts for two representative runs and find the difference negligible (Table~\ref{tab:realized}).

\begin{table}[t]
\centering
\small
\begin{tabular}{lrr}
\toprule
\textbf{Setting} & \textbf{Target} & \textbf{Realized} \\
\midrule
\enar\ Baseline & 0.9\% & {1.3}\% \\
\enar\ WWT & 12.6\% & {13.1}\% \\
\bottomrule
\end{tabular}
\caption{\ceq estimated from target versus realized exposure counts. Using realized counts leaves both the values and their difference essentially unchanged.}
\label{tab:realized}
\end{table}

% % =====================================================================
\section{Intervention Implementation Details}
\label{app:interventions}
% =====================================================================

This appendix provides implementation details and detailed results for the interventions evaluated in \S\ref{sec:pretrainnig_flawed:interventions}.

\subsection{Language Mixing --- Code-switching}
Inspired by lingual code-switching, pretraining code-switching introduces cross-lingual context by randomly swapping words with their cross-lingual counterparts during pretraining. Because the language of the next token becomes stochastic, the model is pushed toward assigning shared representations across languages. We vary two parameters: the \emph{mixing ratio}---the proportion of documents subjected to code-switching---and the \emph{substitution probability}---the probability of replacing any given word within such a document. We execute the substitution using a variant of our curated dictionary presented in (App.~\ref{app:dict}), which skips the conflict resolution stage to avoid the appended IDs on substituted words.

\paragraph{Data exclusion.} We strictly exclude \fkd documents from the mixing pool, simulating the realistic case in which a limited aligned dataset is used alongside a much larger body of standard-format data for which we would like to evaluate transfer.

\paragraph{Results.} We sweep the mixing ratio over $\{0.6, 0.1, 0.01\}$ and the substitution probability over $\{0.5, 0.1, 0.01\}$ (Table~\ref{tab:codeswitch_sweep}). No setting produces meaningful knowledge generalization: \ceq\ ranges from $-1.2\%$ to $2.2\%$ across the sweep, against a baseline of $0.9\%$ (App.~\ref{app:ceq}). There is no clear trend in either parameter --- the best and worst cells differ in mixing ratio by a factor of 60, but the ordering is not monotone in either direction, indicating the differences can be mainly attributed to noise. Code-switching in a natural \enar\ vocabulary-disjoint setting therefore does not appear to induce knowledge sharing at any mixing volume we test.

\begin{table}[t]
\centering
\small
\begin{tabular}{lccc}
\toprule
& \multicolumn{3}{c}{\textbf{Mixing ratio}} \\
\cmidrule(lr){2-4}
\textbf{Subst.\ prob.} & \textbf{0.6} & \textbf{0.1} & \textbf{0.01} \\
\midrule
0.5  & 2.2\% & 0.9\% & $-1.2$\% \\
0.1  & 1.6\% & 0.3\% & 2.0\% \\
0.01 & $-0.4$\% & 1.5\% & 0.4\% \\
\bottomrule
\end{tabular}
\caption{\ceq\ under pretraining code-switching in the \enar\ setting at 360M, across mixing ratio and substitution probability. The native-script baseline is 0.9\%.}
\label{tab:codeswitch_sweep}
\end{table}

\subsection{Activation Alignment Loss}
Activation Alignment loss encourages the model to align inner representations of cross-lingual semantically equivalent units. We extract a global representation of a sequence by concatenating the max and average pooling of its layer-4 activations. We also experiment with multi-depth aggregation of the loss on the SmolLM2-360M architecture, where we average the losses from layers $\{4, 8, 12, 16, 20, 24, 28\}$ (having normalized the losses to account for differences in scales across layers). The auxiliary loss is weighted equally against the cross-entropy objective and applied to 60\%. As with mixing, \fkd documents are excluded from the alignment set.

We evaluate both an InfoNCE contrastive loss \citep{oord2019representationlearningcontrastivepredictive, chi2021infoxlminformationtheoreticframeworkcrosslingual} and an L2 loss. L2 lacks negative sampling and is in principle susceptible to representational collapse; we hypothesise that the primary language-modeling objective provides sufficient regularization to prevent this.

\paragraph{Chunked matches.} Since only document-level parallel data is available across our \enar setup, we utilize an assumption of structural similarity between paired documents --- chunks from similar relative positions in the document carry similar semantics. We choose a unit length $n$, split the language-$A$ document into chunks of $n$ words, and split the language-$B$ document into the same number of equal-length chunks. The choice of $n$ presents a trade-off: large $n$ gives high confidence that paired chunks correspond but poor granularity, while small $n$ gives fine granularity at the risk of misalignment. The structural-similarity assumption is somewhat defensible for machine-translated corpora such as ours due to the structural biases of machine translation.

\paragraph{Results.} We sweep the chunk size $n$ over $\{4, 20, 100, 250\}$ words for both losses (Table~\ref{tab:alignment_sweep}). No configuration offers substantial benefit to knowledge generalization: \ceq\ ranges from 0.3\% to 2.5\% across the sweep, against a 0.9\% baseline.

\begin{table}[t]
\centering
\small
\begin{tabular}{lcccc}
\toprule
& \multicolumn{4}{c}{\textbf{Chunk size $n$ (words)}} \\
\cmidrule(lr){2-5}
\textbf{Loss} & \textbf{4} & \textbf{20} & \textbf{100} & \textbf{250} \\
\midrule
InfoNCE & 2.30\% & 2.22\% & 2.54\% & 2.23\% \\
L2      & 1.89\% & 1.69\% & 0.32\% & 0.47\% \\
\bottomrule
\end{tabular}
\caption{\ceq\ under activation alignment in the \enar\ setting at 360M, across loss type and chunk size. The native-script baseline is 0.9\%.}
\label{tab:alignment_sweep}
\end{table}

% =====================================================================
\section{Soft Mapping}
\label{app:soft_map}
% =====================================================================

Several experiments use \emph{soft mapping}: a single mechanism for partially sharing representations between two token sets, applied in two settings. In this appendix, we define it properly, describe each application, and present detailed results in Tab.~\ref{tab:softmap_sweep}.

\subsection{Mechanism}
Let $d$ be the embedding dimension of our model and $\mathcal{M}$ be a correspondence between two token sets (e.g., $\text{English}_2$ token and its $\text{English}_1$ counterpart). Under soft mapping with sharing rate $p$, each pair $(t, t') \in \mathcal{M}$ shares the first $\lfloor p \cdot d \rfloor$ embedding dimensions as tied parameters updated by gradients from both tokens; the remaining dimensions are independent. Thus $p = 0$ recovers disjoint embeddings and $p = 1$ recovers full mapping. Tokens without a counterpart in $\mathcal{M}$ are unaffected.

\subsection{Diagnostic Use: \texorpdfstring{\enen}{English1-English2}} Here soft mapping is purely a measurement instrument. The two languages are structurally identical and $\mathcal{M}$ is the exact shift correspondence between a token and its copy, so $p$ controls the sole difference between them. Sweeping $p$ answers how much sharing is required before generalization emerges, rather than merely whether it does (\S\ref{sec:tokensRcause:tying}). The sweep also calibrates the upper end of the \ceq scale, since $p \rightarrow 1$ approaches perfect sharing by construction.

\subsection{Practical Use: Soft WWT-Ar}
Under full mapping, WWT-Ar tokens are indistinguishable from their English counterparts. This is what enables knowledge sharing, but it leaves the model no mechanism to represent which language it is generating: the output language must be fixed externally and code-switched generation is unsupported. Soft mapping resolves this---each language retains distinct token identities and shares only embedding dimensions. We do not evaluate code-switching ability in generation directly; we report only that the mechanism exists.

We additionally hypothesized that reserving language-specific dimensions would \emph{improve} transfer, on the grounds that translation equivalents differ in connotation and possible additional meanings and that merging them discards information the model needs. We find no evidence for this: transfer increases with $p$ (\S\ref{sec:TrAr:softmapping}). Soft mapping is therefore best understood as trading a small amount of transfer for the practical properties above, rather than as an improvement over full mapping.

\paragraph{Construction.} Because WWT can change the token count and does not provide a strict one-to-one token correspondence, Soft WWT-Ar requires an additional step: we first perform the WWT while shifting the resulting tokens by $V$ into a disjoint range (as in \enen\ we double the model vocabulary size to accommodate this), then tie a fraction $p$ of each WWT-Ar token's dimensions to the corresponding English token.

\begin{table}[t]
\centering
\small
\begin{tabular}{lrrr}
\toprule
 & & \multicolumn{2}{c}{\textbf{\ceq}} \\
\cmidrule(lr){3-4}
\textbf{$p$} & \textbf{Free dims} & \enen & English--WWT-Ar \\
\midrule
0\%     & {960} & -2.6\%  & --- \\
20\%     & {768} & 8.6\%  & 1.0\% \\
50\%     & {480} & 82.7\% & 1.7\% \\
90\%     & {96}  & 95.8\% & 9.6\% \\
99\%     & {10}  & 99.1\% & 10.5\% \\
99.9\%   & {1}   & 99.0\% & 10.2\% \\
100\%    & 0          & 100\% & 12.6\% \\
\bottomrule
\end{tabular}
\caption{\ceq as a function of the sharing rate $p$ in both settings. \emph{Free dims} is the number of language-specific embedding dimensions remaining at each rate. for \enen\, $p = 1$ corresponds to completely tied language copies --- $\ceq = 100\%$ by definition; for English--WWT-Ar it corresponds to standard (not soft) WWT.}
\label{tab:softmap_sweep}
\end{table}

% =====================================================================
\section{Dictionary Curation and WWT-Ar Construction}
\label{app:dict}
% =====================================================================
This appendix describes how we generate the large-scale word mapping dictionaries used for WWT operation in \S\ref{sec:TrAr:TrAr}. We also provide more information about the costs of this mapping and where they originate.

\subsection{Dictionary Construction}
We parse the first 7M documents of our Arabic corpus and collect all unique words, then translate them with Qwen2.5-32B-Instruct \citep{qwen2025qwen25technicalreport}. The result is a dictionary with 11M entries. Any existing bilingual dictionary could be substituted, avoiding this step entirely. For Russian, we curate a similar dictionary following the same procedure on the Russian corpora. We execute the WWT operation (\S\ref{sec:TrAr:TrAr}) on Arabic and Russian using these dictionaries 

\subsection{Conflict Resolution}
\label{app:dict:conflicts}
The mapping must be reversible. Where multiple Arabic words translate to the same English word, we differentiate them by appending an index: if two Arabic words both map to \textit{dog}, one is remapped to \textit{dog1}. This guarantees that every distinct Arabic word form receives a distinct WWT-Ar form, so no information is lost under the mapping.

Translation is many-to-one at scale --- our dictionaries hold 11.05M Arabic and 13.23M Russian entries mapping onto 6.51M and 6.99M distinct English targets --- but conflicted targets are not uniformly frequent in text, so most replacements need no index. In the Arabic corpus 81.9\% of word replacements take the bare form and 99.7\% need at most a single-digit index; Russian is heavier at 55.8\% and 94.6\% (Table~\ref{tab:conflict_indices}), which we attribute to its richer inflectional morphology, where case and number variants that are distinct surface forms frequently share a single English lemma.

\begin{table}[t]
\centering
\small
\begin{tabular}{lrr}
\toprule
\textbf{Index} & \textbf{Arabic} & \textbf{Russian} \\
\midrule
bare (none) & 81.92\% & 55.78\% \\
1--9        & 17.79\% & 38.82\% \\
10--99      & 0.28\%  & 5.30\% \\
100+        & 0.01\%  & 0.10\% \\
\bottomrule
\end{tabular}
\caption{Share of word replacements taking each conflict-index length.}
\label{tab:conflict_indices}
\end{table}

\subsection{Tokenization Fertility}
\label{app:dict:fertility}
Word-wise mapping does not preserve sequence length. Because dictionary targets and transliterated forms tokenize differently from their source words, mapped documents are longer: WWT-Ar documents average 921 tokens against 709 for native Arabic, an increase of 29.9\%. WWT-Ru averages 1459 tokens against 1190, an increase of 22.6\% (Table~\ref{tab:fertility}).

Conflict resolution indices are the dominant source of WWT's sequence-length overhead, since they turn a word that would tokenize cleanly into a word-plus-digit token sequence. Excluding them, WWT-Ar's fertility penalty falls from $+29.9\%$ to $+17.6\%$ and WWT-Ru's from $+22.6\%$ to $-2.1\%$  (Table~\ref{tab:fertility}). The overhead is therefore a property of our conflict-resolution policy rather than an inherent property of word-wise mapping itself. 

\subsection{Out-of-Dictionary Words} Words absent from the dictionary are handled by reversible transliteration - a character level 1-to-1 mapping into English script. for Arabic we use Buckwalter transliteration and for Russian we use GOST. We measure that in the Arabic corpus 99.7\% of word occurrences are matched rather than transliterated (78\% of unique words), and for Russian that number stands at 99.8\% (92.7\% of unique words).
 
\begin{table}[t]
\centering
\small
\begin{tabular}{lrr}
\toprule
\textbf{Language} & \textbf{Tokens / doc.} & \textbf{$\Delta$} \\
\midrule
Arabic & 709 & --- \\
WWT-Ar & 921 & $+29.9\%$\\
WWT-Ar excluding conflict IDs & 834 & $+17.6\%$ \\
\midrule
Russian & 1190 & --- \\
WWT-Ru & 1459 & $+22.6\%$ \\
WWT-Ru excluding conflict IDs & 1165 & $-2.1\%$ \\
\bottomrule
\end{tabular}
\caption{Mean document length in tokens under the respective \enar\ and \enru\ BPE tokenizers; deltas are reported relative to the native script tokenization. Word-wise mapping increases sequence length, more so for Russian than for Arabic. A substantial part of the fertility increase originates in the appended IDs used for conflict resolution.}
\label{tab:fertility}
\end{table}
 
This carries a genuine inference-time cost: generating a given amount of content under WWT requires proportionally more forward passes than in the native script.
 
It also has a consequence for how our results should be read. Because we hold the token budget fixed rather than the document count, the WWT-Ar and WWT-Ru models see fewer documents than their native-script counterparts---roughly 23\% fewer for WWT-Ar --- and therefore cover less of the foreign-language corpus. The mapped models nonetheless achieve comparable or better English perplexity (\S\ref{sec:TrAr}).
 
One might attribute that gain to reduced interference: with less foreign content competing for capacity, English would benefit. Our results do not support this explanation. English-only pretraining yields \emph{worse} English perplexity than in a bilingual configuration (20.51 against 19.53 and 18.99; \S\ref{sec:pretraining_flawed}), so additional foreign content helps rather than competes at this scale, and the models are not capacity-limited. Reducing Arabic coverage should therefore reduce the benefit English receives, not increase it. That WWT-Ar delivers a larger benefit while contributing fewer documents indicates it contributes more per token---consistent with the increase in knowledge generalization characterized through the \ceq score.
 
% % =====================================================================
\section{Embedding-Space Analysis}
\label{app:embeddings_analysis}
% % =====================================================================

Models fail to generalize knowledge between two identical copies of English (\S\ref{sec:tokensRcause:compart}), yet both copies carry the same language and data distribution, so we should expect the model to learn the same structure for each. To test this, we compare the two halves of each model's embedding matrix and ask whether they are structurally similar, whether one linear map relates them, and whether matching tokens are placed together.

\subsection{Setup}
\label{app:emb:setup}
Corresponding tokens in the two \enen\ vocabularies are directly aligned by row in the embedding matrix. To reduce noise from rarely updated rows, we retain tokens with at least $1{,}000$ expected exposures per language, estimated from corpus frequency and training budget. This leaves $36{,}170$ rows at 360M and $37{,}062$ at 7B, out of $65{,}536$ per vocabulary. To prevent a shared mean direction from inflating similarity, we normalize rows, mean-center each matrix, and renormalize. 

We report four measures: \emph{matched cosine}, the mean cosine between corresponding rows; \emph{linear CKA} \citep{kornblith2019similarity} for global structure; \emph{mutual $k$-NN} \citep{huh2024platonic} at $k=10$ for local structure; and \emph{linear-map cosine}, the matched cosine on a held-out $20\%$ after fitting a least-squares map on the remaining $80\%$.

Within-model comparisons use all eligible rows at each scale; across-model comparisons use the shared eligible set. We analyze the tied embedding matrix at 360M and the LM head at 7B.\footnote{The 360M architecture ties its input embedding and output projection; the 7B architecture separates them. We use the 7B LM head because it receives the vocabulary-wide prediction updates that also train the tied 360M matrix, whereas its input embedding remains close to initialization.} Following \citet{groger2026revisiting}, we calibrate each measure against $K=200$ random row permutations at $\alpha=0.05$. On the resulting $0$--$1$ scale, higher means more similar: $0$ marks the permutation baseline and $1$ identical spaces. Every within-model score exceeds all permutations ($p=0.005$).

\begin{table}[H]
\centering
\small
\setlength{\tabcolsep}{4pt}
\begin{tabular}{lccccc}
\toprule
\textbf{Intervention} & \ceq & \textbf{Matched} & \textbf{Linear} & \textbf{Mutual} & \textbf{Mapped} \\
& & \textbf{cosine} & \textbf{CKA} & \textbf{$k$-NN} & \textbf{cosine} \\
\midrule
\multicolumn{6}{l}{\emph{360M}} \\
Baseline & $-2.6\%$ & 0.012 & 0.682 & 0.654 & 0.560 \\
Code-switching (60\%) & $71.8\%$ & 0.691 & 0.841 & 0.740 & 0.755 \\
Same initialization & $78.6\%$ & 0.939 & 0.957 & 0.854 & 0.960 \\
\midrule
\multicolumn{6}{l}{\emph{7B}} \\
Baseline & $-1.5\%$ & 0.013 & 0.611 & 0.819 & 0.506 \\
\bottomrule
\end{tabular}
\caption{Null-calibrated embedding similarity and \ceq\ across \enen\ interventions. The baselines learn similar, linearly alignable spaces but show negligible direct token correspondence and no knowledge transfer; code-switching and same initialization bridge the spaces and transfer knowledge.}
\label{tab:embeddings}
\end{table}

\subsection{Results}
\label{app:emb:results}
\paragraph{Within models.}
At 360M, the baseline already has strong global and local structural similarity (Table~\ref{tab:embeddings}). Direct token correspondence is nevertheless negligible (matched cosine $0.012$), while mapped cosine reaches $0.560$ on held-out tokens. A single linear map therefore recovers substantial correspondence absent from the raw space. The 7B baseline repeats this pattern.

We further analyze the embeddings of two models that achieve high knowledge transfer. The model trained with the same initialization for corresponding tokens (\S\ref{sec:tokensRcause:drift}) and substantial code-switching in 60\% of the documents, which transfers well in the \enen\ setting, unlike the more realistic \enar\, (\S\ref{sec:pretrainnig_flawed:interventions}). In both models, we observe a sharp rise in matched cosine similarity alongside the jump in \ceq\ (Table~\ref{tab:embeddings}). Global and local similarities also increase from already strong baselines, indicating that the main change is direct token alignment.

\begin{figure}[H]
\centering
\includegraphics[width=\linewidth]{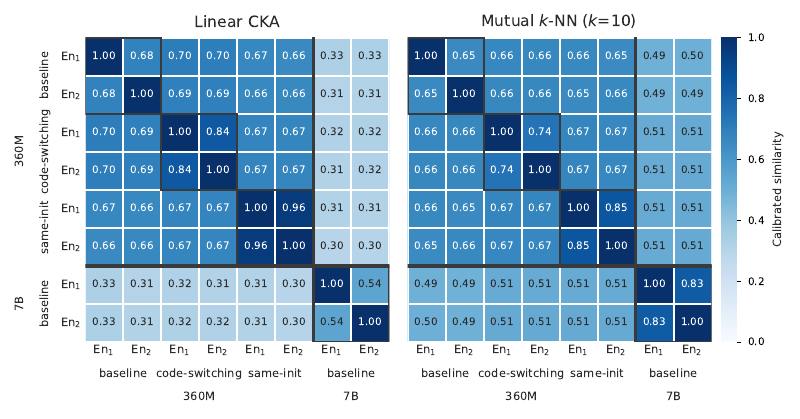}
\caption{Pairwise null-calibrated global (linear CKA; left) and local (mutual $k$-NN; right) embedding similarity. All models learn a common English geometry; code-switching and same initialization additionally bridge its two tokenized copies, yielding higher within-model similarity.}
\label{fig:embedding_similarity}
\end{figure}

\paragraph{Across models.}
At 360M, the baseline's two halves fall within the same CKA and mutual $k$-NN ranges as cross-model pairs (Figure~\ref{fig:embedding_similarity}). The baseline is thus no more structurally similar within one run than across runs: all 360M spaces learn a common English geometry. Models that generalize add a stronger within-model relation. Across scales, global and local similarities are lower than within-scale comparisons but remain substantial and well above the permutation baseline, despite differences in width and training budget. Combined with high matched cosine in the transfer conditions (Table~\ref{tab:embeddings}), this shows that the interventions bridge corresponding tokens within a common geometry; shared geometry alone does not yield knowledge transfer.

% =====================================================================
\section{Token Overlap and Anchored Arabic}
\label{app:AnAr}
% =====================================================================

Languages that share a writing system naturally exhibit partial token overlap through shared subwords, and this overlap correlates with better cross-lingual generalization \citep{ifergan2024beneathsurfaceconsistencyexploring, kallini2025falsefriendsfoesinvestigating}. This appendix examines whether recreating that overlap artificially induces knowledge sharing. It supports two claims in the main text: that the effectiveness of shared tokens depends on their frequency, not merely their number (\S\ref{sec:tokensRcause:tying}), and that partial token-space overlap between natural languages does not deliver the gains of full merging (\S\ref{sec:TrAr:res}).

\subsection{Controlled Overlap for \texorpdfstring{\enen}{English1-English2}}
\label{app:AnAr:controlled}
We first sweep token overlap in \enen, where a subset of tokens is shared outright between the two copies while the remainder stay disjoint. Generalization rises steeply with the shared subset: sharing the 2\% most frequent tokens already yields $\ceq = 59.9\%$, reaching 89.0\% at 64\% overlap (Table~\ref{tab:overlap}). Sharing a subset of tokens completely is thus far more effective than sharing all tokens partially --- 2\% overlap outperforms 20\% soft-mapping by a wide margin (App.~\ref{app:soft_map}).

Vocabulary share alone, however, does not determine the outcome. We repeat the 8\% condition with a different token subset --- one covering 47\% of running text rather than 69.1\% --- and \ceq\ falls from 66.9\% to 51.7\%. The two conditions share the same fraction of the vocabulary and differ only in how frequently the shared tokens appear, so generalization tracks the \emph{coverage} of the shared tokens rather than their count. (This second subset is the anchor set of the AnAr mapping introduced in \S\ref{app:AnAr:anar}.)

\subsection{Anchored Arabic}
\label{app:AnAr:anar}
\begin{wrapfigure}{r}{0.6\textwidth} % "r" aligns right, reserving 55% of the page width
    \centering
    \includegraphics[width=0.6\columnwidth]{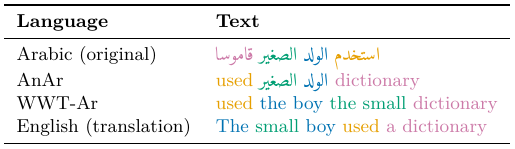}
    \caption{An Arabic sentence under each representation. AnAr replaces only words with a direct token-to-token dictionary match, leaving the rest in Arabic script; WWT-Ar maps every word. Both preserve Arabic word order (notice native Arabic is written from right to left) and grammar, unlike the English translation.}
    \label{fig:mapping_comparison}
\end{wrapfigure}

Following these results we construct AnAr, a mapping that replaces only those Arabic tokens with a direct token-to-token match in our dictionary, leaving the rest in Arabic script (Figure~\ref{fig:mapping_comparison}). AnAr targets the same barrier as WWT-Ar but far less aggressively, producing partial token overlap rather than a fully merged token space. Practically, AnAr preserves sequence length, avoiding the fertility overhead of WWT-Ar.

AnAr nonetheless fails to deliver meaningful knowledge generalization, with $\ceq = 1.6\%$ against the WWT-Ar ceiling of 12.6\%. The controlled sweep explains why: AnAr's anchors comprise 8\% of the vocabulary but cover only 47\% of English and 34.6\% of Arabic running text, this low coverage, as we established, weakens the effect of partial vocabulary sharing (App.~\ref{app:AnAr:controlled}), partially explaining the gap between effectiveness here and in the \enen setup, more linguistic aspects may be in play.

\begin{table}[t]
\centering
\small
\begin{tabular}{llr}
\toprule
\textbf{Overlap} & \textbf{Coverage} & \textbf{\ceq} \\
\midrule
\multicolumn{3}{l}{\emph{$\text{English}_1-\text{English}_2$}} \\
0\% & 0\% & $-2.6$\% \\
0.1\% & 11.4\% & $-1.3$\% \\
0.5\% & 29.0\% & 43.1\% \\
2\% & 50.1\% & 59.9\% \\
4\% & 58.5\% & 61.2\% \\
\textbf{8\%} & \textbf{69.1\%} & \textbf{66.9\%} \\
\textbf{8\%} & \textbf{47\%} & \textbf{51.7\%} \\
16\% & 79.2\% & 68.2\% \\
32\% & 88.5\% & 77.2\% \\
64\% & 96.1\% & 89.0\% \\
\midrule
\multicolumn{3}{l}{\emph{$\text{English}-\text{Arabic}$}} \\
0\% & 0\% & 0.9\% \\
\multicolumn{3}{l}{\emph{$\text{English}-\text{AnAr}$}} \\
8\% & 47\%/34.6\% & 1.6\% \\
\multicolumn{3}{l}{\emph{$\text{English}-\text{WWT-Ar}$}} \\
100\% & 100\% & 12.6\% \\
\bottomrule
\end{tabular}
\caption{Knowledge generalization as a function of token overlap. \emph{Overlap} is the share of vocabulary types shared between the two languages, and \emph{coverage} is the share of training tokens that they account for, reported per side where the two differ. The controlled setting achieves far higher \ceq than the natural one, while within each setting \ceq increases with overlap. These experiments were performed at the 360M parameter scale.}
\label{tab:overlap}
\end{table}

% =====================================================================
\section{Language Modeling under WWT}
\label{app:ar_lm}
% =====================================================================

The main text reports that WWT improves English perplexity (\S\ref{sec:TrAr:res}). This appendix provides the corresponding non-English measurements, so that the gain can be read against what it costs on the mapped side, and reports the monolingual controls that isolate how much each representation extracts from the English half of the corpus (Tab.~\ref{tab:bpb}).

\subsection{Measurement}
\label{app:lm:metric}
We report bits per byte (based on Arabic script byte count) --- BPB --- rather than perplexity. WWT-Ar and native-script Arabic differ in token fertility (App.~\ref{app:dict:fertility}), and a per-token measure would mechanically favour the higher-fertility representation: each WWT-Ar token carries less of the underlying text and is correspondingly easier to predict.

\subsection{Arabic-Side Modeling}
\label{app:lm:arabic}
In the bilingual setting, mapping Arabic to WWT-Ar costs little on the Arabic side: at 360M parameters bits per byte rises from $0.7129$	 to $0.7160$, a relative degredation of less the 0.5\%. similarly, at 7B parameter scale BPB goesfrom $0.5679$ to $0.5741$, degradation of $1.1\%$. This is a small cost for the $14\times$ improvement in \ceq\ reported in \S\ref{sec:TrAr:res}, and it is achieved while the WWT-Ar model is exposed to roughly 23\% fewer Arabic documents under our fixed token budget (\S\ref{sec:TrAr:TrAr}).

\subsection{Monolingual Controls}
\label{app:lm:monolingual}
Bilingual numbers alone cannot say how much of each model's Arabic-side performance comes from the English half of the corpus. We therefore train monolingual counterparts for both representations and compare each bilingual model against its monolingual baseline (Table~\ref{tab:bpb}), trained only on the Arabic half of the data (30\% less data for WWT-Ar). The quantity of interest is the \emph{improvement} from adding English: if the unified token space allows the model to draw more from English, WWT-Ar should show the larger improvement. Indeed, we find that the monolingual Arabic model outperforms the WWT-Ar model with BPB of $0.7155$ compared to $0.7344$ (2.6\% difference), however, the WWT-Ar model gains much more from the cross-lingual data --- a $\Delta$ of 0.0184 compared to 0.0026, roughly $7\times$ matching directional \ceq measurement gain (Tab.~\ref{tab:bpb}).

\begin{table}[t]
\centering
\small
\begin{tabular}{lccc}
\toprule
\textbf{Representation} & \textbf{Monolingual} & \textbf{Bilingual} & $\Delta$ \\
\midrule
% \multicolumn{3}{l}{\emph{360M Parameters}} \\
Native Arabic & 0.7155 & 0.7129 & 0.0026 \\
WWT-Ar        & 0.7344 & 0.7160 & 0.0184 \\
\midrule
Native Russian & 0.5892 & 0.5763 & 0.0129 \\
WWT-Ru         & 0.6164 & 0.5870 & 0.0294 \\
\bottomrule
\end{tabular}
\caption{Non-English language modeling in bits per byte (BPB), normalized by native-script bytes throughout (\S\ref{app:lm:metric}); lower is better. \emph{Bilingual} models are trained on a 50/50 token split with English, \emph{monolingual} models on the non-English half alone. $\Delta$ is the improvement contributed by the English half. Experiments performed at the 360M parameter scale.}
\label{tab:bpb}
\end{table}

% % =====================================================================
% \section{English--Russian Replication}
% \label{app:ru}
% % =====================================================================
% % TODO(NEW)

% We replicate our core compartmentalization result on a second language pair, pretraining on English and Russian using \textit{FineWeb2-HQ} \citep{messmer2026enhancingmultilingualllmpretraining} for the Russian side.

% \todo{Report setup differences from the English/Arabic experiments --- corpus,
% tokenizer, data volume, whether the 50/50 token split is maintained.}
% \todo{Report baseline \ceq, and TrRu if that experiment was run.}
% were not.}

% =====================================================================
\section{Shuffled-Mapping Controls}
\label{app:shuffle}
% =====================================================================

Our WWT operation is overwhelmingly a translation rather than a transliteration operation (\S\ref{sec:TrAr:TrAr}), meaning it maps Arabic tokens or token sequences to their semantically matching English counterparts. This appendix provides implementation details for the experiments performed in \S\ref{sec:TrAr:meanings_matter} to determine the importance of this property as opposed to plain token superposition. We run three controls that preserve token sharing while destroying correspondence, and find they all fail. We perform these experiments at the 360M parameter scale.

\subsection{Shuffled WWT Dictionary}
We destroy semantic correspondences by shuffling the keys and targets of our large-scale dictionaries used for WWT, then proceeding with the WWT process as usual, mapping words to their shuffled dictionary mapping and transliterating out-of-vocabulary entries. We find this process hurts knowledge generalization between the two languages compared to our unshuffled WWT baseline, with English--WWT-Ar \ceq falling from 12.6\% to 3\%, and English--WWT-Ru falling from 23.3\% to 3.6\%.

\subsection{Transliteration}
Transliteration-only mapping provides a similar effect: a unified token space, but tokens aren't mapped to their semantic counterparts. We perform the same reversible transliteration used for WWT but apply it to all words. Essentially applying WWT with an empty dictionary. We find that transliterating Arabic does not help with resolving compartmentalization between it and English, with a \ceq score of $-0.3\%$.

\subsection{Permuted \texorpdfstring{\enen}{English1-English2}}
\label{app:permuted_enen}
We additionally pretrain a model on two language clones --- \enen\ --- in which both languages occupy a single shared token space, but the token identities of one copy are permuted: $\sigma: \{0, \dots, V-1\} \to \{0, \dots, V-1\}$. Formally, while a text sequence is tokenized in English\textsubscript{1} as an array of token IDs $(t_1, t_2, \dots, t_n)$, English\textsubscript{2} tokenizes the parallel sequence into $(\sigma(t_1), \sigma(t_2), \dots, \sigma(t_n))$, both share the same  $\{0, \dots, V-1\}$ token space

This provides a fully controlled environment where the entire token inventory is shared and the two languages remain structurally identical, with the only barrier to generalization being the mismatch between token idetities. Yet, we observe no knowledge generalization with a \ceq score of -2.6\% --- same as completely disjoint token spaces.

\subsection{Interpretation}
Shared tokens transfer knowledge only when they carry corresponding meanings. Resolving disjoint tokenization is therefore not a matter of token superposition, and a semantically faithful mapping is necessary. The transliteration control is notable, as previous work has suggested transliteration as a path to unlocking multilingual ability \citep{moosa-etal-2023-transliteration, j-etal-2024-romansetu, saji-etal-2025-romanlens, jayakumar2026scriptstimesurveyevolving}, we note that successful examples in past literature focus mostly on related languages where cognates are common and therefore transliteration can supply the semantic matches we find necessary.

\begin{table}[t]
\centering
\small
\begin{tabular}{lccr}
\toprule
\textbf{Setting} & \textbf{Token space} & \textbf{Semantic mapping} & \textbf{\ceq} \\
\midrule
\multicolumn{4}{l}{\emph{$\text{English}-\text{Arabic}$}} \\
Native Arabic (baseline)   & disjoint & ---  & 0.9\% \\
WWT-Ar                     & shared   & yes  & 12.6\% \\
Shuffled WWT-Ar            & shared   & no   & 3.0\% \\
Transliterated Arabic      & shared   & no   & $-0.3$\% \\
\midrule
\multicolumn{4}{l}{\emph{$\text{English}-\text{Russian}$}} \\
Native Russian (baseline)  & disjoint & ---  & 1.9\% \\
WWT-Ru                     & shared   & yes  & 23.3\% \\
Shuffled WWT-Ru            & shared   & no   & 3.6\% \\
\midrule
\multicolumn{4}{l}{\emph{$\text{English}_1-\text{English}_2$}} \\
Disjoint copies (baseline) & disjoint & ---  & $-2.6$\% \\
Fully shared copies        & shared   & yes  & 100\% \\
Permuted copies            & shared   & no   & $-2.6$\% \\
\bottomrule
\end{tabular}
\caption{Semantic controls at 360M. Each block pairs a semantically
faithful shared token space against controls that share tokens without
sharing meanings. Sharing a token inventory without semantic
correspondence performs at or near the disjoint-token baseline in every
pair, while the semantically faithful mapping is the only condition
that improves generalization. for \enen\ Fully shared copies are at \ceq of 100\% by definition}
\label{tab:shuffle_controls}
\end{table}

\end{document}